\documentclass[10pt,twocolumn,letterpaper]{article}

\usepackage[pagenumbers]{style/cvpr}

\usepackage{booktabs}
\usepackage{multirow}
\usepackage{multicol}
\usepackage{enumitem}

\setitemize{noitemsep,topsep=0pt,parsep=0pt,partopsep=0pt}

\usepackage[dvipsnames]{xcolor}
\usepackage{circledsteps}
\usepackage{bbding}
\usepackage{makecell}
\usepackage{anyfontsize}
\usepackage{tcolorbox}
\usepackage{listings}
\definecolor{codegreen}{rgb}{0,0.6,0}
\definecolor{codegray}{rgb}{0.5,0.5,0.5}
\definecolor{codepurple}{rgb}{0.58,0,0.82}
\definecolor{backcolour}{rgb}{0.95,0.95,0.95}

\lstdefinestyle{mystyle}{
    backgroundcolor=\color{backcolour},   
    commentstyle=\color{codegreen},
    keywordstyle=\color{codepurple},
    numberstyle=\tiny\color{codegray},
    stringstyle=\color{codepurple},
    basicstyle=\ttfamily\footnotesize,
    breakatwhitespace=false,         
    breaklines=true,                 
    captionpos=b,                    
    keepspaces=true,                 
    numbers=left,                    
    numbersep=5pt,                  
    showspaces=false,                
    showstringspaces=false,
    showtabs=false,                  
    tabsize=2,
}

\lstdefinelanguage{PyTorch}{%
  language     = Python,
  morekeywords = {ones, T, where, argmax, update, shape, get_min_concept, log_and_keep},
}

\newcommand{\rcnt}[1]{\rotatebox[origin=c]{90}{#1}}

\newcommand{\eric}[1]{\textcolor{NavyBlue}{#1}}
\newcommand{\es}[1]{\eric{#1}}

\newcommand{\xhdr}[1]{\vspace{4pt}\noindent\textbf{#1}}
\newcommand{\xhdrflat}[1]{\noindent\textbf{#1}}

\newcommand{\cpos}[1]{\textcolor[HTML]{73CF9B}{#1}}
\newcommand{\cneg}[1]{\textcolor[HTML]{E37C74}{#1}}
\newcommand{\cneu}[1]{\textcolor[HTML]{646464}{#1}}

\newcommand{\cmagenta}[1]{\textcolor[HTML]{E63888}{#1}}
\newcommand{\cgreen}[1]{\textcolor[HTML]{009C3B}{#1}}
\newcommand{\cblue}[1]{\textcolor[HTML]{2799F6}{#1}}
\newcommand{\cgray}[1]{\textcolor[HTML]{858585}{#1}}

\newcommand{\myquote}[1]{``\emph{#1}''}

\newcommand{\fairdedup}{FairDeDup\xspace}

\usepackage{paralist}
\usepackage[dvipsnames]{xcolor}

\definecolor{cvprblue}{rgb}{0.21,0.49,0.74}
\usepackage[pagebackref,breaklinks,colorlinks,citecolor=cvprblue]{hyperref}

\usepackage[capitalize]{cleveref}
\crefname{section}{Sec.}{Secs.}
\Crefname{section}{Section}{Sections}
\crefname{table}{Tab.}{Tabs.}
\Crefname{table}{Table}{Tables}

\def\paperID{8758}
\def\confName{CVPR}
\def\confYear{2024} 

\title{FairDeDup: Detecting and Mitigating Vision-Language\\ Fairness Disparities in Semantic Dataset Deduplication}

\author{Eric Slyman$^{1,2}$\thanks{Work conducted during Slyman's 2023 summer internship at Adobe.} \quad Stefan Lee$^{1}$ \quad Scott Cohen$^{2}$ \quad Kushal Kafle$^{2}$ \vspace{0.3em} \\
{\normalsize $^1$Department of EECS, Oregon State University} \quad
{\normalsize $^2$Adobe}
}

\begin{document}
\maketitle

\begin{abstract}
Recent dataset deduplication techniques have demonstrated that content-aware dataset pruning can dramatically reduce the cost of training Vision-Language Pretrained (VLP) models without significant performance losses compared to training on the original dataset. These results have been based on pruning commonly used image-caption datasets collected from the web -- datasets that are known to harbor harmful social biases that may then be codified in trained models. In this work, we evaluate how deduplication affects the prevalence of these biases in the resulting trained models and introduce an easy-to-implement modification to the recent SemDeDup algorithm that can reduce the negative effects that we observe. When examining CLIP-style models trained on deduplicated variants of LAION-400M, we find our proposed \fairdedup algorithm consistently leads to improved fairness metrics over SemDeDup on the FairFace and FACET datasets while maintaining zero-shot performance on CLIP benchmarks.
\end{abstract}    

\section{Introduction}
\label{sec:introduction}
\begin{figure}[t]
    \centering
    \includegraphics[width=\linewidth]{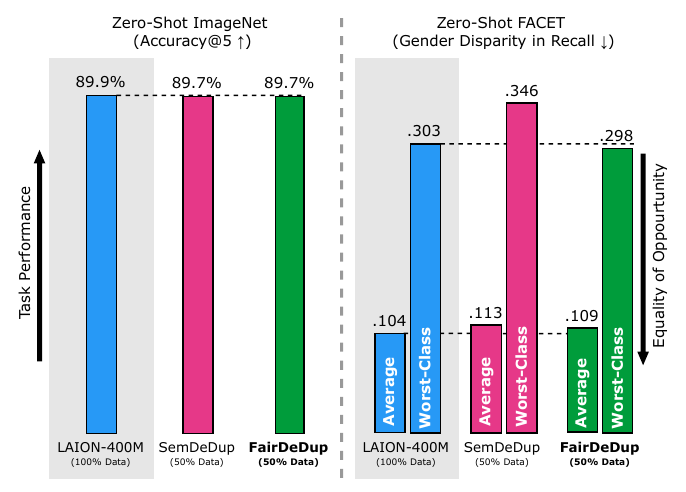}
    \caption{Training models on deduplicated data can yield similar results to the full-data setting on standard tasks like zero-shot ImageNet~\cite{deng2009imagenet} classification (\textbf{left}, higher is better $\uparrow$). However, impacts on subgroup performance have not been studied. We discover cases such as gender disparity (\textbf{right}, lower is better $\downarrow$) where deduplication reinforces existing biases on FACET~\cite{gustafson2023facet}. \fairdedup preserves performance while reducing bias from deduplication and, in some cases, w.r.t. the full-data setting.}
    \label{fig:teaser}
\end{figure}

Recent Vision-Language Pretrained (VLP) models \cite{radford2021clip} that learn to align image and language encodings have demonstrated strong zero-shot performance on many standard perception tasks \cite{cherti2023clipbenchmark,deng2009imagenet,young2014flickr,zhai2019vtab}. Beyond these, VLP models have enabled complex downstream applications ranging from visually-aware chatbots \cite{liu2023improved,li2023blip} and language-based image segmentation \cite{kirillov2023segment,zou2023segment} to instruction-guided robotics \cite{xiao2022skill,shridhar2022cliport} and semantic mapping of 3D scenes \cite{shafiullah2022clip, kerr2023lerf}. 
The rapid adoption and widespread impact of these models is due in part to the incredibly broad range of content they can represent effectively -- a scope far exceeding prior models trained on manually-curated, closed-world datasets \cite{deng2009imagenet,lin2014coco}. To acquire this capability, VLP models are trained on massive open-world datasets of image-caption pairs collected from the internet \cite{schuhmann2021laion}. VLP models improve reliably with additional training data \cite{cherti2023reproducible}, driving the number of examples in these datasets into the billions \cite{schuhmann2022laion}. This scale of uncurated data introduces at least two challenges -- 1) training can be extremely costly, and 2) manual data curation to reduce undesirable social biases is economically prohibitive. In this work, we explore how dataset deduplication techniques developed to reduce training costs may exacerbate or ameliorate these biases in trained models.

While larger pretraining datasets generally yield better model performance \cite{cherti2023reproducible}, the massive web-scraped datasets commonly used for training VLP models contain many identical samples (duplicates) or samples that capture nearly the same content under similar imaging conditions (semantic duplicates \cite{abbas2023semdedup}). Several recently developed techniques for data pruning/deduplication have demonstrated that aggressive removal of these duplicates has limited impact on the task performance of trained models \cite{sorscher2022beyond,abbas2023semdedup,maharana2023d2pruning}. For example, \citet{abbas2023semdedup} found that pruning LAION-400M~\cite{schuhmann2021laion} by 50\% resulted in trained models that achieved average performance within $0.5$\% of their full-data analogs across a range of common benchmark tasks -- effectively cutting training time in half.

However, these web-scale datasets contain a plethora of problematic social biases and harmful stereotypes \cite{birhane2021multimodal,garcia2023phase,birhane2023into}. These biases can often then be reflected in the behavior of models trained on these datasets \cite{agarwal2021evaluating,hendricks2016women,wang2022measuring,hirota2022quantifying}. To better understand and reduce these potential harms, there is increased interest in analyzing the composition of these datasets and their downstream effects on trained models \cite{garcia2023phase,birhane2021multimodal,Zhao2017MenAL}. Deduplication techniques introduce another algorithmic step between dataset and model training that may systematically alter the data distribution -- potentially amplifying, maintaining, or reducing the effect of dataset biases. Given that deduplication techniques will likely be widely deployed as cost-saving measures, understanding how their design affects the behavior of downstream models in terms of bias and fairness is a timely but unexamined question.

To study this question, we investigate the fairness outcomes of CLIP-style \cite{radford2021clip} VLP models trained on the LAION-400M dataset~\cite{schuhmann2021laion} pruned with SemDeDup~\cite{abbas2023semdedup}. Replicating the results of \citet{abbas2023semdedup}, we find task performance on CLIP Benchmark \cite{cherti2023clipbenchmark} is only marginally affected; however, evaluation on the fairness-focused FairFace \cite{karkkainen2021fairface} and FACET \cite{gustafson2023facet} datasets suggest deduplication results in mixed effects compared to the full-data setting. We observe increased disparities across gender, but both positive and negative changes for disparities across skin tone and age. Based on these findings, we propose \fairdedup\ -- a fairness-aware data pruning algorithm that makes pruning decisions to improve representation of specified \emph{sensitive concepts} (\eg, \emph{gender}, shown in \cref{fig:teaser}). The implementation of \fairdedup is a simple modification to SemDeDup and specifying concepts can be done in natural language. Our large-scale experiments show that \fairdedup leads to improved fairness outcomes comapred to SemDeDup while maintaining comparable performance on standard zero-shot and retrieval-based performance benchmarks. To better understand the deduplication process, we run a smaller scale study deduplicating demographic-labeled data -- finding that \fairdedup consistently retains more images depicting minority classes than SemDeDup.

\xhdr{Contributions.} We summarize our contributions below:
\begin{compactitem}[\hspace{-2pt}•]
    \item We conduct, to our knowledge, the first large-scale experiment evaluating the fairness outcomes of training large-scale vision language models on pruned data -- training CLIP-style models on full and deduplicated versions of the popular LAION-400M dataset then evaluating on standard fairness benchmarks for VLP models.
    \item We find that models trained on SemDeDup \cite{abbas2023semdedup} pruned data have varied effects on fairness outcomes from the full-data model; reinforcing some biases and mitigating others.
    \item We introduce \fairdedup, a simple and efficient modification to SemDeDup that improves fairness outcomes while retaining task performance -- improving fairness outcomes over SemDeDup in nearly all cases studied. 
\end{compactitem}

\section{Related Work}
\label{sec:related}

\xhdrflat{Vision-Language Fairness.}
Vision and language models have been shown to learn, reflect, and amplify problematic social biases. For example, vision systems have been shown to dehumanize minority groups by identifying them as animals \cite{nyt2015google} and degrade in task performance on intersectional combinations of gender and skin tone~\cite{buolamwini2018gendershades}. Likewise, language models are known to learn gendered associations of professions~\cite{bolukbasi2016debiasing}, increase sentiment-intensity along racial lines~\cite{kiritchenko2018examining}, and a myriad of other problems documented in \cite{blodgett2020languagebias,sun2019genderbias}. Vision-language models are not exempt from these problems \cite{nyt2021facebook,hendricks2016women,noble2018algorithms} and can even reinforce them \cite{srinivasan2021worst, Zhao2017MenAL}.

Contemporary Vision-Language Pretrained models are frequently pretrained on massive but uncurated data scraped from the internet \cite{radford2021clip,chen2023pali,jia2021align,li2021albef}. While web-scale data is shown to improve performance, it also teaches models \myquote{misogyny, pornography, and malignant stereotypes} \cite{birhane2021multimodal}. VLP models demonstrate dehumanizing behavior with respect to racial subgroups in zero-shot text-image retrieval~\cite{agarwal2021evaluating,berg2022prompt}, show bias related to gender \cite{hendricks2016women,wang2022measuring,hirota2022quantifying,garcia2023phase}, age~\cite{garcia2023phase} and skin tone \cite{zhao2021understanding,wang2022measuring,hirota2022quantifying,garcia2023phase} in image captioning, and also demonstrate biases relating to age, gender, skin tone, and ethnicity in text-image retrieval~\cite{garcia2023phase,zhou2022vlstereoset}. These behaviors are attributed to the use of uncurated web-scale datasets in pretraining VLP models~\cite{birhane2021multimodal,garcia2023phase,birhane2023into}.

Mitigations for bias in VLP models typically include fairness-aware training~\cite{zhang2022counterfactually} or post-hoc methods to disentangle useful concepts from sensitive attributes~\cite{berg2022prompt,seth2023dear,chuang2023debiasing}. Unlike these methods, we seek to prevent bias from being reinforced in the dataset, rather than removing bias from the model itself. Though early vision-language fairness literature frequently calculates WEAT~\cite{caliskan2017weat} and SEAT~\cite{may2019seat} embedding association measures extended for the multimodal setting~\cite{ross2020measuring,janghorbani2023multimodal}, these measures have been shown to be overly sensitive to small changes in model architecture and outputs~\cite{berg2022prompt}. As such, VLP model fairness is primarily evaluated on CelebA~\cite{liu2015celeb} and FairFace~\cite{karkkainen2021fairface}. Recent datasets such as PHASE~\cite{garcia2023phase} and FACET~\cite{gustafson2023facet} allow for the study of bias on ``in the wild'' data across diverse subgroups. 

\xhdr{Dataset Pruning.}
Several techniques exist for reducing the size of a dataset while preserving, or even improving, performance. We consider all techniques under this umbrella as \emph{dataset pruning} algorithms. Coreset selection chooses a weighted subset of training samples which closely estimate the full dataset's gradient~\cite{mirzasoleiman2020coresets,guo2022deepcore} to perform data-efficient training with little loss in performance. However, these methods do not scale well with dataset size and frequently require class labels~\cite{sorscher2022beyond}. The most similar work to ours among coreset selection algorithms is the recent D2 Pruning~\cite{maharana2023d2pruning}. D2 Pruning utilizes graph based methods to select samples that are both \emph{hard} and \emph{diverse} across a data distribution. While promising, D2 Pruning does not evaluate any fairness outcomes and is only demonstrated to scale to DataComp Small (12.8M)~\cite{gadre2023datacomp}, a low accuracy setting for VLP models peaking around $5\%$ top-1 zero-shot ImageNet~\cite{deng2009imagenet} accuracy. In comparison, base-sized CLIP-style models can range from $67$-$74\%$ accuracy with web-scale data on the same task. We refer readers to~\cite{phillips2016coresets} for a more thorough review of coreset selection algorithms. 

Large-scale deduplication typically attempts to find exact perceptual duplicates using techniques like perceptual hashing~\cite{du2020perceptual} or filtering~\cite{gadre2023datacomp} on image-text CLIP scores and target classes (\eg, filtering to images close to ImageNet classes). \citet{abbas2023semdedup} introduces the concept of semantic duplicates, images with similar semantic meaning that are not perceptually the same image, alongside SemDeDup, a formalized version of the unsupervised deduplication algorithm from \cite{sorscher2022beyond}. SemDeDup has been shown to be capable of significantly reducing dataset size with only marginal impact on performance. We choose to study SemDeDup due to the ubiquity of its underlying selection method among contemporary deduplication algorithms - cosine similarity between samples - and scalable nature. 
To our knowledge, we are the first to study the effect of data pruning on the fairness outcomes of VLP models and study the effects of fairness-aware pruning on their behavior.

\section{\fairdedup: Fair Semantic Deduplication}
\label{sec:methodology}
\begin{figure}[t]
     \centering
     \includegraphics[width=0.98\linewidth]{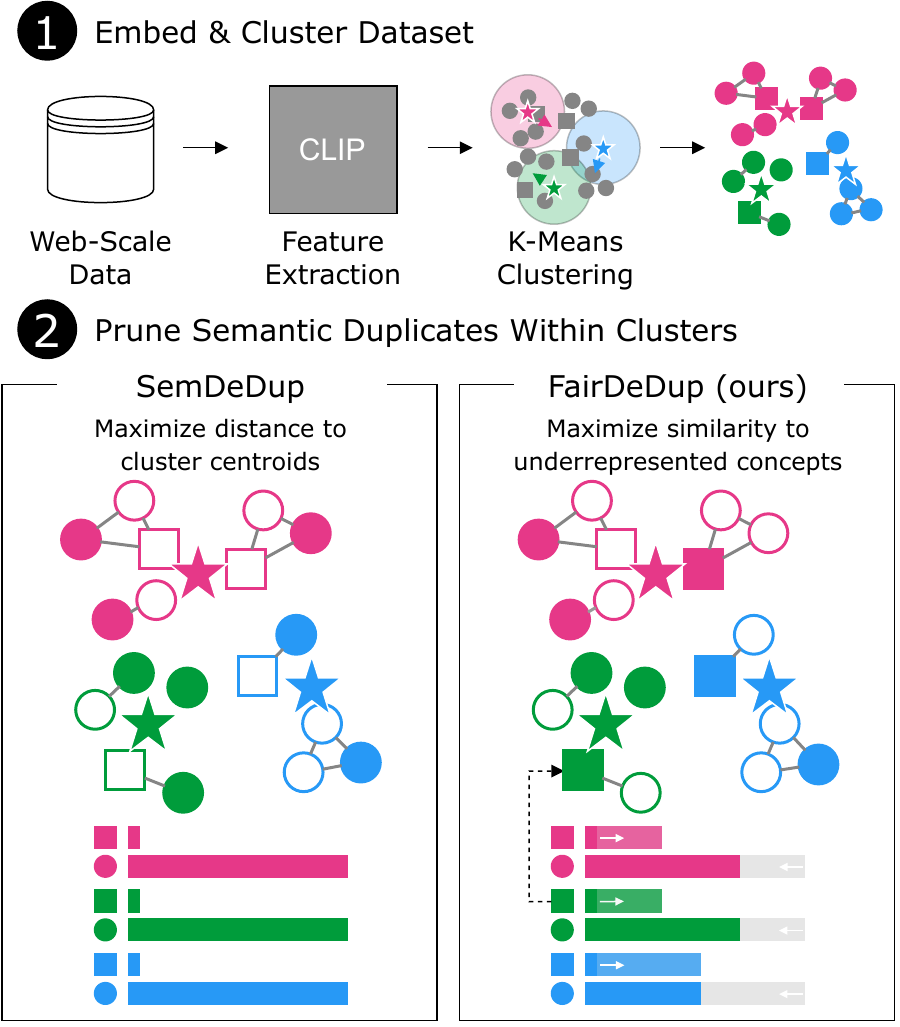}
     \caption{The semantic deduplication pipeline following three clusters (\cmagenta{\tiny\FiveStar},\cgreen{\tiny\FiveStar},\cblue{\tiny\FiveStar}) with two subgroups (\cgray{\tiny\SquareSolid},\cgray{\tiny\CircleSolid}). Connected shapes are duplicates. We \textbf{(1) embed} all images from the dataset with a pretrained model then partition with $k$-means to enable efficient search during \textbf{(2) deduplication}. We make a simple modification to the maximum distance selection heuristic used by \citet{abbas2023semdedup} \textbf{(left)} to improve subgroup diversity by preserving samples which maximize similarity to poorly represented sensitive concepts according to user-specified concept prototypes (\textbf{right}).}
     \label{fig:overview}
\end{figure}

There frequently exists sensitive attributes in data for which it is desirable to obtain some notation of fairness \cite{franklin2022ontology}. For example, we may seek \emph{demographic parity} for \emph{gender} so that individuals do not receive differing treatment based on their gender identity. Such outcomes are usually based on social norms, organizational ethics, or even codified into discrimination law~\cite{ieee2020ethics,biden2023executive,uslaw1964titlevii,eu2008discrimination}. Our goal is to improve post-deduplication fairness outcomes concerning these sensitive groups. To achieve this, we propose boosting the representation of underrepresented sensitive subgroups on the internet (\eg, women of color) in the post-pruning dataset distribution. We allow for user-defined natural language \emph{sensitive concepts}, which captures these subgroups for consideration in the deduplication process, and leverage them to bias the selection of preserved samples towards those concepts which are currently underrepresented.

\subsection{Preliminaries: SemDeDup}
We implement FairDeDup as a lightweight modification to the SemDeDup algorithm, which we describe here for completeness. \citet{abbas2023semdedup} identify that pruning both exact perceptual duplicates (\eg, copies of the same image) and those that carry redundant semantic information (\eg, many photos of the same object from differing angles), denoted semantic duplicates, is helpful for improving the data efficiency of training large models. To achieve this, they propose SemDeDup~\cite{abbas2023semdedup}, an extension of the unsupervised pruning metric from \citet{sorscher2022beyond} to web-scale data.

To identify duplicates, SemDeDup first leverages pre-trained foundation models (\eg CLIP~\cite{radford2021clip}) to embed all images in the dataset into a semantically meaningful feature space. Na\"ively thresholding embedding similarity between all points to detect duplicates requires $\mathcal{O}(n^2)$ pairwise comparisons and is intractable for web-scale data like LAION-400M, which requires computing ${\approx}1.5{\times}10^{17}$ cosine similarities. To mitigate this, the dataset is partitioned using an efficient $K$-means algorithm under the assumption that pairwise similarity need only be calculated for approximately similar samples. SemDeDup then considers the resulting $\mathcal{O}(n^2/k)$ pairwise similarities on an independent per cluster basis. Within each cluster, they determine sets of samples within a $1{-}\epsilon$ similarity threshold as duplicates and keep only the sample most distant from the cluster centroid. While this selection heuristic is motivated by the hardness hypothesis of \citet{sorscher2022beyond}, ablations show that the algorithm is robust to choosing even a random sample.

\subsection{FairDeDup}
Due to the robustness of SemDeDup to the choice of selection heuristic on performance, we seek instead to replace the heuristic with one that can support our fairness motivation. We provide an overview following shared and unique steps of SemDeDup and \fairdedup in \cref{fig:overview}.

\xhdr{Sensitive Concept Prototypes.}
Given a list of user-defined sensitive concepts $C$ that are desired to be represented in the pruned dataset, we denote the \emph{concept prototype} $P_i$ for a sensitive concept $C_i{\in}C$ as the average text embedding of the set of captions generated from template strings (\eg, \myquote{A photo of a \{$C_i$\}}) capturing that concept. 
As is common for VLP models, we assume the embedding model supporting image clustering can also produce image-text alignment scores~\citep{radford2021clip,li2022blip,zhang2021vinvl,jia2021align,yu2022coca} and consider the case where alignment is determined as the cosine similarity between the representations produced by a vision $\Phi_I: I \rightarrow \mathbb{R}^d$ and text $\Phi_T: T \rightarrow \mathbb{R}^d$ encoder: 
\begin{equation}
    sim(I, T) = \Phi_I(I)^T\Phi_T(T)\ /\ \lVert\Phi_I(I)\rVert \lVert\Phi_T(T)\rVert.
\end{equation}
We measure how well an image aligns with a sensitive concept by measuring the image-text similarity between that image and the concept prototype $sim(I, P_i)$. We choose concepts that both relate to commonly protected demographic subgroups of people and are annotated in common fairness datasets, such as ones based in race and gender. Additional details on the selection of sensitive concepts (\cref{sec:sup-concepts}) and a list of all concepts used (\cref{sec:sup-prototype}) are given in the appendix. While this work focuses on text-based prototypes, we note that our methodology trivially extends to image-based ones and beyond, as described in \cref{sec:discussion}.

\xhdr{Sample Preservation Heuristic.} 
To determine which samples to prune, we consider duplicate \emph{neighborhoods}: the set of images within $1{-}\epsilon$ similarity of a given point, and preserve only one sample from each neighborhood. For each cluster produced by $k$-means, we track the running average similarity between preserved samples in that cluster and the sensitive concept prototypes. Until all samples are visited, we randomly select an unvisited sample, calculate the similarity between all samples in its neighborhood and the prototypes, and keep only the sample that maximizes similarity to the least similar running average prototype, marking all points in the neighborhood as visited. We preserve the sample with the highest average similarity across concept prototypes for the first neighborhood visited in a cluster.

We track running average similarity on a per cluster basis
for two reasons: 1) to avoid a synchronous update step between workers processing clusters in parallel and 2) 
to prevent algorithmic ``gaming'' of the selection criteria by balancing concept representation on clusters which highly represent a concept due to some stereotyped notion. Given two clusters primarily composed of doctors and nurses, for example, per cluster processing prevents balancing underselection of feminine presenting doctors by overselecting feminine presenting nurses. We provide pseudo-code for the \fairdedup selection heuristic in \cref{fig:pseudocode}. We visualize random samples after pruning a cluster manually identified to be primarily composed of people with \fairdedup and the SemDeDup maximum distance selection heuristic in \cref{fig:select-doctor}, and show additional examples in the appendix (\cref{sec:sup-selection}).

\begin{figure}[t]
\centering
\begin{lstlisting}[language=PyTorch,escapechar=\%]
# Input: prototypes, embeddings, eps
# Get similarity with concept prototypes
proto = embeddings @ prototypes.T

balance = AverageMeter(prototype.shape[0])
tovisit = torch.ones(embeddings.shape[0])
while tovist.any():
    # Find an unvisited neighborhood
    node = torch.where(tovisit)[0][0]
    sims = embeddings[node] @ embeddings.T
    neighbors = torch.where(sims > 1 - eps)[0]

    # Maximize least represented concept
    c = balance.get_min_concept()
    point = proto[neigbors][:, c].argmax()
    balance.update(point)

    log_and_keep(point)
    tovisit[neighbors] = 0
\end{lstlisting}
\caption{PyTorch-style pseudo-code for \fairdedup selection given concept \texttt{prototypes}, within cluster \texttt{embeddings}, and an \texttt{eps} similarity threshold for determining neighborhoods. We omit the base case where the first sample selected within a cluster is the one with the highest average concept prototype similarity.}
\label{fig:pseudocode}
\end{figure}
\begin{figure}[t]
    \centering
    \begin{subfigure}[b]{0.95\linewidth}
         \centering
         \includegraphics[width=\linewidth]{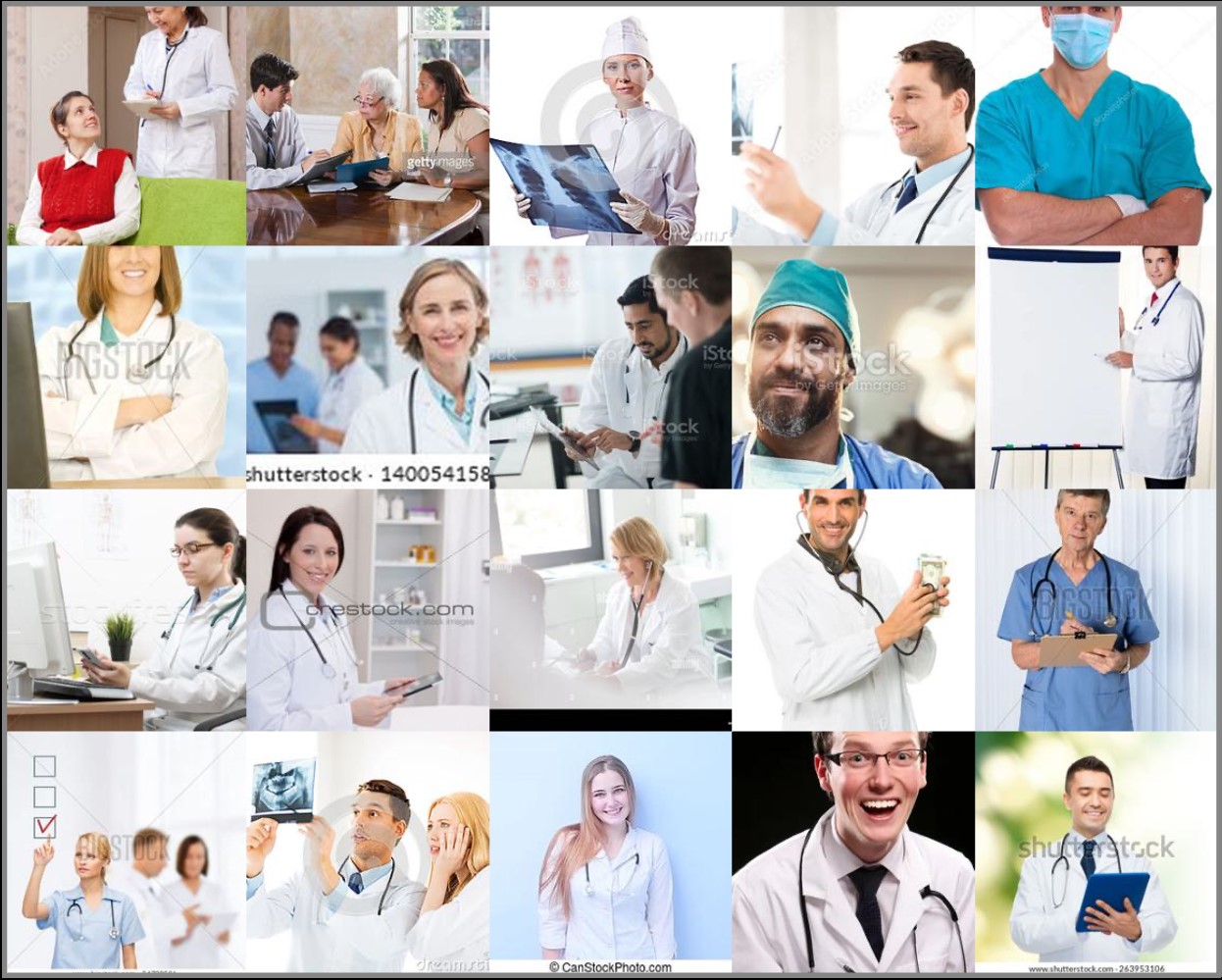}
         \caption{Maximum Distance Selection}
         \label{fig:select-doctor-sdd}
     \end{subfigure}
    \\
    \begin{subfigure}[b]{0.95\linewidth}
         \centering
         \includegraphics[width=\linewidth]{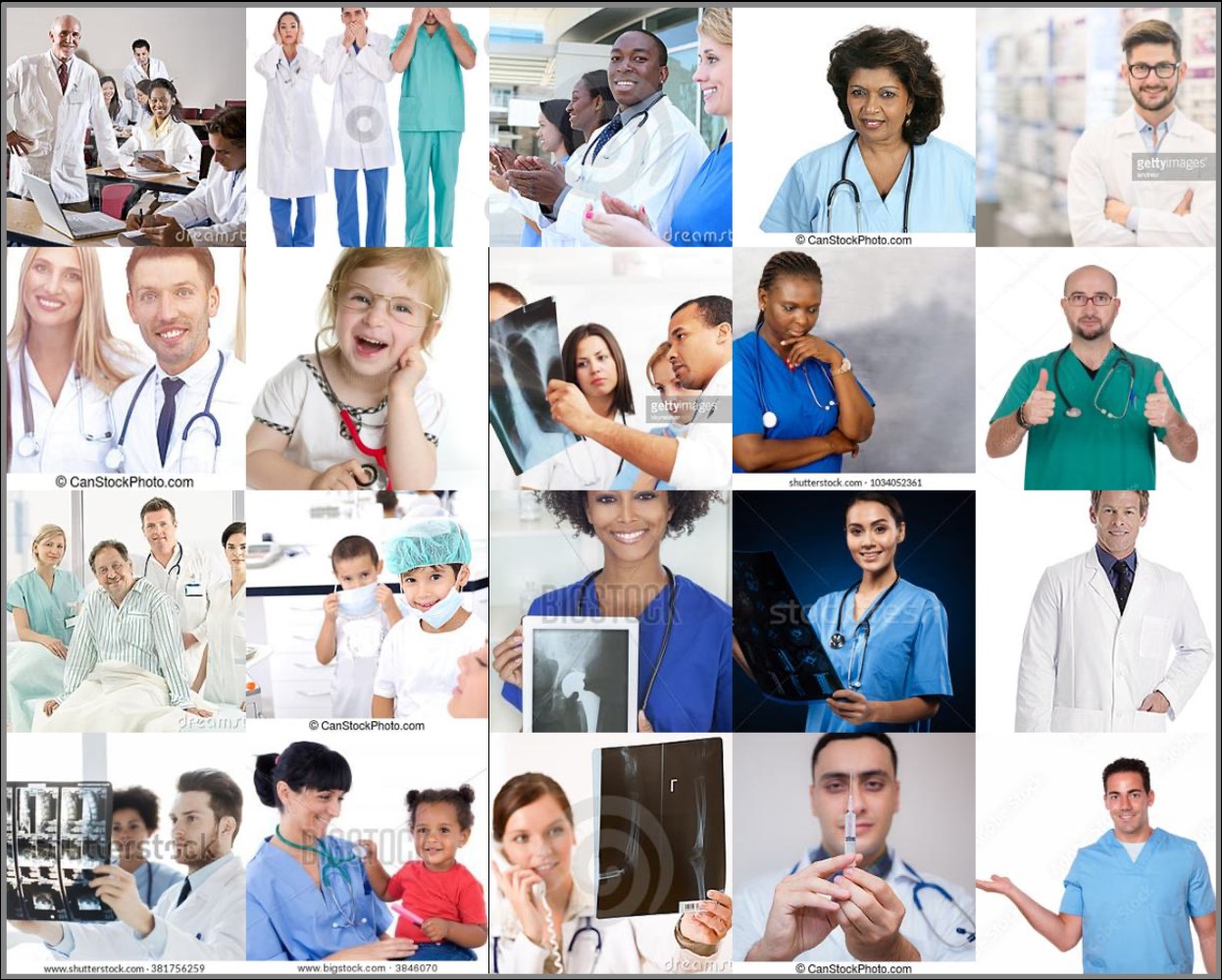}
         \caption{\fairdedup Selection}
         \label{fig:select-doctor-fdd}
     \end{subfigure}
     \caption{A random sampling of preserved samples from a cluster primarily composed of medical professionals after deduplication. \fairdedup improves selection diversity featuring increased variability in age, skin tone, and gender presentation.}
     \label{fig:select-doctor}
\end{figure}

\section{Experiments}
\label{sec:experiments}

To assess the effect of deduplication on learned VLP models, we train CLIP-style models on variants of LAION-400M~\cite{schuhmann2021laion} and evaluate their performance on both standard and fairness-oriented benchmarks for zero-shot classification and text-image retrieval.

\subsection{Models and Training }
We train all models on LAION-400M~\cite{schuhmann2021laion} as a web-scale dataset representative of those typically used for large-scale vision-language pretraining. LAION-400M contains image-text pairs extracted from Common Crawl\footnote{\href{https://commoncrawl.org/}{https://commoncrawl.org/}} filtered to have image-text CLIP similarity ${\geq}0.3$ without significant further curating. This makes LAION-400M an ideal test case for our setting as it is sufficiently large to train VLP models, captures bias from the internet, and is expected to contain semantically redundant samples. At the time of our data collection, only 375M image-text pairs from LAION-400M were still available for download.

We train CLIP-ViT-Base/16~\cite{radford2021clip} models from the OpenClip~\cite{ilharco2021openclip} implementation with vision transformer~\cite{dosovitskiy2021vit} base (ViT-B-16) as the image encoder and text transformer~\cite{vaswani2017attention} as the text encoder. We perform distributed training over 80-120 A100 GPUs depending on the model with a global batch size of 33,820 image-caption pairs for 16 epochs regardless of dataset size. We use the AdamW~\cite{loshchilov2017adamw} optimizer with linear warm up and cosine annealed learning rate schedule peaking at $5{\times}10^{-4}$. Additional hyperparameter details are provided in the appendix (\cref{sec:sup-hyperparameters}).

We evaluate CLIP training on three LAION-400M data settings for \emph{performance} and \emph{fairness}:

\xhdr{Baseline: LAION-400M.} 
We train a CLIP model on the full LAION-400M dataset for a total of 183k steps as a control by which to evaluate baseline performance and fairness. A good model in the deduplicated setting should perform similarly to this model on common benchmarks without negatively impacting subgroup disparity and skew.

\xhdr{SemDeDup LAION-400M.}
For SemDeDup, we use a CLIP-ViT-Base/16 trained on WebImageText (WIT)~\cite{radford2021clip} to produce image embedding which are partitioned into $50,000$ clusters using the FAISS~\cite{johnson2019faiss} implementation of $k$-means and set the $\epsilon$ threshold for identifying duplicates within each cluster such that $50\%$ of samples are pruned.

\xhdr{\fairdedup LAION-400M.}
We leverage the same WIT trained CLIP model for \fairdedup as SemDeDup. We consider 110 sensitive concepts capturing intersectional combinations of age, gender, skin tone, race and ethnicity and represent them using embeddings of 330 corresponding captions (three each with minor syntactic variation). We use the average across captions of the same concept as our prototypes. We enumerate all sensitive concepts and templates used to generate the text prototypes in the appendix (\cref{sec:sup-prototype}). The selection step can be parallelized across CPUs up to the number of clusters produced by $k$-means. We find that selection in this setting on a 32 CPU machine takes one hour on average and that the overall time is dominated by the shared GPU parallelizable embedding and clustering steps.

\subsection{Datasets and Metrics}
We evaluate across three benchmarks to validate if models trained on deduplicated data are both \emph{performant} and \emph{fair}.

\xhdr{Zero-Shot Classification and Retrieval.}
We evaluate the performance of each model across 41 common zero-shot classification and retrieval datasets from Clip~Benchmark~\cite{cherti2023clipbenchmark} such as ImageNet~\cite{deng2009imagenet}, Flickr30k~\cite{young2014flickr}, and VTAB~\cite{zhai2019vtab}. A model trained on deduplicated data should perform at least as well as a model trained in the full-data setting on these benchmarks.

\xhdr{Fair Zero-Shot Classification.}
The FACET~\cite{gustafson2023facet} dataset contains expert reviewer annotations for 52 person related classes, gender presentation, skin tone, age, and other attributes, on a 32k image subset of Segement Anything 1 Billion (SA-1B)~\cite{kirillov2023segany}. We perform zero-shot classification over the 52 person-classes by constructing a text prompt (\eg, \myquote{A photo of a \{class\}}) for each class and predicting the class used to construct the prompt with highest similarity to the image. Given a model $f$, sensitive attribute label $l$, person-class $\mathcal{C}$, and set of of images $\mathcal{I}_{l}^{C}$ which captures class $\mathcal{C}$ featuring a person with label $l$, we measure the average and worst-class disparity in recall between subgroups of sensitive attributes where disparity is defined as:
\begin{equation}
\begin{split}
    disparity = recall(f&(l_1,\mathcal{I}_{l_1}^{C},\mathcal{C})) \\
    &- recall(f(l_2,\mathcal{I}_{l_2}^{C},\mathcal{C})).
\end{split}
\label{eq:disparity}
\end{equation}
Conceptually, a large magnitude disparity indicates that a model better predicts positive instances of person-class $\mathcal{C}$ for one of the two subgroups, while a disparity of zero indicates \emph{equality of opportunity} between subgroups.

We evaluate subgroup disparity for average perceived gender expression by masculine \vs feminine presentation, lighter (1-4MST\footnote{Monk Skin Tone scale \cite{monk2019skin}}) \vs darker (6-10MST) skin tone, and middle \vs younger and middle \vs older age for all person-classes which have at least 25 samples in both subgroups. \citet{gustafson2023facet} consider only the 21k images capturing a single person in their disparity analysis for simplicity and alignment between tasks (\eg, classification and visual grounding). To increase the sample size of our analysis, we consider each person in the dataset as a unique sample.  We expand the bounding box for each person by 20\% to capture context before extracting a square-padded image crop centered on the box, yielding 49,551 images.

\xhdr{Fair Image Retrieval.}
FairFace~\cite{karkkainen2021fairface} annotates a balanced dataset of 108k cropped faces from YFCC-100M~\cite{thomee2016yfcc} by seven racial groups with additional annotations for perceived gender and age. Similar to \cite{berg2022prompt,seth2023dear,chuang2023debiasing}, we measure the degree to which the top-$k$ results of an image-text query differ over values of sensitive attributes in the 11k image validation set with respect to the desired proportion of those values with MaxSkew@1000~\cite{geyik2019fairness}. Given the top-$k$ images $\tau_r^k$ returned by image-text query $r$, let the actual proportion of images returned by the query for a particular value $a_i{\in}A$ of sensitive attribute $A$ be $P_{\tau_r^k,r,a_i}{\in}[0,1]$ and the desired proportion be $P_{q,r,a_i}{\in}[0,1]$, then the skew of value $a_i$ is:
\begin{equation}
    Skew_{a_i}@k(\tau_r) = \ln\left(\frac{P_{\tau_r^k,r,a_i}}{P_{q,r,a_i}}\right)
\label{eq:skew}
\end{equation}
One limitation of Skew@k is that it is defined only for a single value of a sensitive attribute. To give a more holistic view across all values that a sensitive attribute may take on, we report the most skewed $a_i$ with MaxSkew@k:
\begin{equation}
    MaxSkew@k(\tau_r) = \max_{a_i\in A}Skew_{a_i}@k(\tau_r)
\label{eq:maxskew}
\end{equation}
Conceptually, MaxSkew indicates the \myquote{largest
unfair advantage}~\cite{geyik2019fairness} provided to images with a particular value of the sensitive attribute for appearing in the the top-$k$ results of the query. We choose the desired proportion of images to be the same as the true distribution of those images in the dataset. Under this condition, if the proportion of of $a_i$ in the top-$k$ results is the same as its distribution in the dataset, MaxSkew obtains an optimal result of $0$ and achieves \emph{demographic parity.} Following \cite{berg2022prompt}, we report average MaxSkew@1000 across 240 (un)favorable captions orthogonal to images in the dataset (\eg, \myquote{A photo of a \{smart\} person}), matching test attributes and prompts for race ($|A|{=}7$), gender ($|A|{=}2$), and age ($|A|{=}3$). Similar to \citet{seth2023dear}, we bin age into larger groups: \emph{younger} (0-19), \emph{middle} (20-49), and \emph{older} (50-70+) to reduce noise.

We additionally report MinSkew@k, which captures the \myquote{worst disadvantage in representation} for a subgroup, and the normalized discounted cumulative KL-divergence (NDKL), which captures the weighted average of Skew@k over all attribute values at varying settings of $k$. Intuitively, MinSkew captures the severity of the most negatively biased subgroup juxtaposed against the most positively biased captured by MaxSkew, and NDKL is a summary statistic over configurations of Skew@k. We refer readers to \citet{geyik2019fairness} for the formulation of MinSkew and NDKL.

\section{Results}
\label{sec:results}
\begin{table}[t]
\centering
\footnotesize
\begin{tabular}{rccc}
\toprule
& \makecell{\textbf{Full Data} \\[-5pt] {\tiny ($100\%$)}}
& \makecell{\textbf{SemDeDup}  \\[-5pt] {\tiny ($50\%$)}}
& \makecell{\textbf{\fairdedup} \\[-5pt] {\tiny ($50\%$)}}
\\ \midrule
IN1K$_{acc@5}$    & {.899} & {.897 \cneg{(--.002)}} & {.897 \cneg{(--.002)}} \\
INV2$_{acc@5}$    & {.845} & {.841 \cneg{(--.004)}} & {.837 \cneg{(--.008)}} \\
C10$_{acc@5}$     & {.999} & {.998 \cneg{(--.001)}} & {.999 \cneu{(--.000)}} \\
C100$_{acc@5}$    & {.934} & {.934 \cneu{(--.000)}} & {.939 \cpos{(+.005)}}  \\
Flickr$_{R@5}$    & {.873} & {.874 \cpos{(+.001)}}  & {.871 \cneg{(--.002)}} \\
COCO$_{R@5}$      & {.633} & {.632 \cneg{(--.001)}} & {.626 \cneg{(--.007)}} \\ \bottomrule
\end{tabular}
\caption{Common zero-shot and text-image retrieval benchmarks for CLIP models on ImageNet1K~\cite{deng2009imagenet}, ImageNetV2~\cite{recht2019imagenetv2}, CIFAR~\cite{krizhevsky2009cifar} (C10/C100), Flicker30k~\cite{young2014flickr}, and COCO Captions~\cite{lin2014coco}. Higher ($\uparrow$) is better in all cases. The difference in performance from the full-data setting is shown in \cpos{green} (\cneg{red}) when improved (reduced). Both deduplication strategies yield models that preserve the performance of models trained on the full data.}
\label{tab:benchmark}
\end{table}

\xhdrflat{Deduplication Preserves Aggregate Performance.}
In \cref{tab:benchmark}, we report Accuracy@5 for four common zero-shot image classification datasets: ImageNet1K~\cite{deng2009imagenet}, ImageNetV2~\cite{recht2019imagenetv2}, CIFAR-10 and CIFAR-100~\cite{krizhevsky2009cifar}, and Recall@5 for two common image-text retrieval datasets: Flicker30k~\cite{young2014flickr} and COCO Captions~\cite{lin2014coco}. As expected, the performance drop from the full-data setting to deduplicated is marginal (${\leq}0.8\%$), indicating that performance is preserved after pruning $50\%$ of the training data. The performance gap between the two deduplicated-data models is even smaller (${\leq}0.6\%$), and neither consistently performs more favorably across tasks. We refer readers to \cref{sec:sup-benchmarks} in the appendix for results on additional datasets and metrics.

\xhdr{SemDeDup Has Mixed Effects on Fairness.}
We show the result of our zero-shot image classification evaluation on FACET~\cite{gustafson2023facet} in \cref{tab:facet}, studying subgroups across gender, skin tone, and age-based sensitive attributes. We find that SemDeDup yields mixed impacts. SemDeDup reinforces average and worst-class disparity across gender subgroups, exacerbates average disparity in skin tone while mitigating the worst-class, and surprisingly aids in reducing average and worst-class disparity across age groups.

In \cref{tab:fairface}, we present the results of our text-image retrieval evaluation on FairFace~\cite{karkkainen2021fairface}, focusing on subgroups related to gender, race, and age. We again find that SemDeDup demonstrates mixed effects. SemDeDup reinforces gender skew across all metrics but mitigates skew towards the largest unfairly advantaged group (MaxSkew) while magnifying skew away from the worst disadvantaged group.
\begin{table}[t]
\centering
\footnotesize
\begin{tabular}{l@{\hspace{.25em}}lrcccc}
\toprule
& &
& \makecell{\textbf{Full} \\[-5pt] {\tiny ($100\%$)}}
& \makecell{\textbf{SemDeDup}  \\[-5pt] {\tiny ($50\%$)}}
& \makecell{\textbf{FairDeDup} \\[-5pt] {\tiny ($50\%$)}}
& \makecell{\textbf{Diff.} \\[-5pt] {\tiny (FDD-SDD)}}
\\ \midrule
\multirow{3}{*}{\rcnt{\bf Gender}} & \multirow{3}{*}{\rcnt{\tiny Masc / Femm}}  
  & \footnotesize{Mean} & .104 & .113 \cneg{(+\hphantom{0}9\%)} & {\bf.109 \cneg{(+\hphantom{0}5\%)}}  & --.004 \\
& & \footnotesize{Max}  & .303 & .346 \cneg{(+14\%)} & {\bf.298 \cpos{(--\hphantom{0}2\%)}} & --.048 \\
& & \footnotesize{Gap}  & .199 & .233 \cneg{(+17\%)} & {\bf.189 \cpos{(--\hphantom{0}5\%)}} & --.053 \\
\midrule
\multirow{3}{*}{\rcnt{\bf Skin Tone}} & \multirow{3}{*}{\rcnt{\tiny Light / Dark}} 
  & \footnotesize{Mean} & .100 & .112 \cneg{(+12\%)}  & {\bf.105 \cneg{(+\hphantom{0}5\%)}}   & --.007 \\
& & \footnotesize{Max}  & .354 & .342 \cpos{(--\hphantom{0}3\%)} & {\bf.320 \cpos{(--10\%)}}  & --.022 \\
& & \footnotesize{Gap}  & .254 & .230 \cpos{(--\hphantom{0}9\%)} & {\bf.215 \cpos{(--15\%)}}  & --.015 \\
\midrule
\multirow{3}{*}{\rcnt{\bf Age}} & \multirow{3}{*}{\rcnt{\tiny Mid / Yng}}    
  & \footnotesize{Mean} & .063  & {\bf.059 \cpos{(--\hphantom{0}6\%)}} & .075 \cneg{(+19\%)} & +.016 \\
& & \footnotesize{Max}  & .268  & .230 \cpos{(--14\%)}      & {\bf.225 \cpos{(--16\%)}} & --.005 \\
& & \footnotesize{Gap}  & .205  & .171 \cpos{(--17\%)}      & {\bf.150 \cpos{(--27\%)}} & --.022 \\
\midrule
\multirow{3}{*}{\rcnt{\bf Age}} & \multirow{3}{*}{\rcnt{\tiny Mid / Old}}
  & \footnotesize{Mean} & .098  & .096 \cpos{(--\hphantom{0}2\%)} & {\bf.087 \cpos{(--11\%)}} & --.009 \\
& & \footnotesize{Max}  & .252  & .248 \cpos{(--\hphantom{0}2\%)} & {\bf.153 \cpos{(--39\%)}} & --.095 \\
& & \footnotesize{Gap}  & .154  & .152 \cpos{(--\hphantom{0}1\%)} & {\bf.066 \cpos{(--57\%)}} & --.008 \\
\bottomrule
\end{tabular}
\caption{\es{Absolute d}isparity (Eq.~\ref{eq:disparity}) in zero-shot classification performance  on FACET~\cite{gustafson2023facet} \es{averaged} across 52 person classes. Larger values indicate a greater performance gap between subgroups when predicting true positive samples of the same occupation. Lower ($\downarrow$) is better for all metrics. Best deduplicated model in \textbf{bold}. The percent change in fairness outcomes from the full-data setting is shown in \cpos{green} (\cneg{red}) when improved (reduced).}
\label{tab:facet}
\end{table} 

\xhdr{\fairdedup Improves Fairness Over SemDeDup.}
\fairdedup improves fairness outcomes over SemDeDup on FACET by mitigating, rather than exasperating, worst-class gender disparity while improving disparity outcomes in all cases except for age between middle-aged and young subgroups. With respect to SemDeDup, \fairdedup reduces the average over groups for mean disparity by $.0001$ ($.0067$ excluding \texttt{Age Mid/Yng}), worst-class by $.0425$ and gap by $.0245$. This result demonstrates that \fairdedup more closely achieves \emph{equality of opportunity} than SemDeDup.

FairFace also shows evidence that \fairdedup improves fairness outcomes. While both methods increase gender skew, \fairdedup exhibits a milder skew across all summary metrics. For race, both methods mitigate the effects of the largest unfairly advantaged group (MaxSkew) compared to the baseline, while \fairdedup mitigates the magnitude of MaxSkew and reduces the skew against the worst disadvantaged class (MinSkew) compared to the baseline. Determining the best-performing method for age-based subgroups is inconclusive. Across gender and race groups, \fairdedup reduces MinSkew by $.0555$, MaxSkew by $.0285$ and NDKL by $.0035$. This results demonstrates that \fairdedup better achieves \emph{demographic parity} than SemDeDup w.r.t. gender and race, even outperforming the full-data setting on race.
\begin{table}[t]
\centering
\footnotesize
\newcommand{\bftab}{\fontseries{b}\selectfont}
\begin{tabular}{lrcccc}
\toprule
&
& \makecell{\textbf{Full} \\[-5pt] {\tiny ($100\%$)}}
& \makecell{\textbf{SemDeDup}  \\[-5pt] {\tiny ($50\%$)}}
& \makecell{\textbf{\fairdedup} \\[-5pt] {\tiny ($50\%$)}}
& \makecell{\textbf{Diff.} \\[-5pt] {\tiny (FDD-SDD)}}
\\ \midrule
\multirow{3}{*}{\rcnt{\bf Gender}}
& \footnotesize{MinSkew} & .159 & .223 \cneg{(+40\%)} & {\bf.182 \cneg{(+14\%)}} & --.041 \\
& \footnotesize{MaxSkew} & .123 & .153 \cneg{(+24\%)} & {\bf.125 \cneg{(+\hphantom{0}2\%)}} & --.028 \\
& \footnotesize{NDKL}    & .010 & .015 \cneg{(+50\%)} & {\bf.012 \cneg{(+20\%)}} & --.003 \\
\midrule
\multirow{3}{*}{\rcnt{\bf Race}}
& \footnotesize{MinSkew} & .545 & .583 \cneg{(+\hphantom{0}7\%)}  & {\bf.513 \cpos{(--\hphantom{0}6\%)}} & --.070 \\
& \footnotesize{MaxSkew} & .432 & .401 \cpos{(--\hphantom{0}7\%)} & {\bf.372 \cpos{(--14\%)}} & --.029 \\
& \footnotesize{NDKL}    & .035 & .034 \cpos{(--\hphantom{0}3\%)} & {\bf.030 \cpos{(--14\%)}} & --.004 \\
\midrule
\multirow{3}{*}{\rcnt{\bf Age}}
& \footnotesize{MinSkew} & .618 & .702 \cneg{(+14\%)}  & {\bf.647 \cneg{(+\hphantom{0}5\%)}} & --.055 \\
& \footnotesize{MaxSkew} & .241 & {\bf.224 \cpos{(--\hphantom{0}7\%)}} & .296 \cneg{(+23\%)} & +.072 \\
& \footnotesize{NDKL}    & .023 & {\bf.022 \cpos{(--\hphantom{0}4\%)}} & .028 \cneg{(+22\%)} & +.006 \\
\bottomrule
\end{tabular}
\caption{Skew evaluation on FairFace~\cite{karkkainen2021fairface} averaged over 240 text-image query templates. As MinSkew is a negative metric optimal at its upper bound of zero, we report its absolute value for readability so that lower ($\downarrow$) is better for all metrics. Best deduplicated model in \textbf{bold}. The percent change in fairness outcomes from the full-data setting is shown in \cpos{green} (\cneg{red}) when improved (reduced).}
\label{tab:fairface}
\end{table}

\section{Discussion}
\label{sec:discussion}

Below we discuss observations when pruning smaller-scale annotated data, potential \fairdedup variants for varied concept prototypes, and limitations of our approach.

\xhdr{Evaluation on Demographically Annotated Data.}
In this paper, we have shown on large-scale real model training that \fairdedup achieves results on-par with SemDeDup on standard benchmarks, while demonstrating improved fairness outcomes. We believe that is the clearest signal about the  applicability of \fairdedup in real-world usage. However, we would also like to directly demonstrate that \fairdedup does indeed select more diverse data representations compared to SemDeDup. To do so, we consider deduplicating the FACET~\cite{gustafson2023facet} images described in ~\cref{sec:experiments}. We perform $k$-means clustering ($k{=}50$) on the images with ten different random seeds and apply both deduplication methods to each. In \cref{tab:facet-distribution}, we report the percent of the post-pruning dataset labeled as non-majority classes for gender (\emph{feminine}, \emph{non-binary}, \emph{other}), skin tone (\emph{$MST{>}4$}, \emph{other}), and age (\emph{younger}, \emph{older}, \emph{other}), averaged across the ten trials. This analysis indicates that 1) SemDeDup does indeed reduce the frequency of the least well represented sub-groups and 2) that \fairdedup mitigates this effect. The difference between means across trials of SemDeDup and \fairdedup is statistically significant at ${\geq}99.9\%$ confidence ($n{=}10$) across all groups according to a paired t-test.

\begin{table}[t]
\centering
\begin{tabular}{lccc}
\toprule
& \makecell{\textbf{Full Data} \\[-5pt] {\tiny ($100\%$)}}
& \makecell{\textbf{SemDeDup}  \\[-5pt] {\tiny ($50\%$)}}
& \makecell{\textbf{\fairdedup} \\[-5pt] {\tiny ($50\%$)}}
\\ \midrule
{Gender}    & {32.92\%} & {31.91\%} & {\bf 32.29\%} \\ 
{Skin Tone} & {51.28\%} & {50.46\%} & {\bf 51.06\%} \\
{Age}       & {44.74\%} & {43.62\%} & {\bf 44.06\%} \\ \bottomrule 
\end{tabular}
\caption{Data mass allocated to minority classes in FACET~\cite{gustafson2023facet} after deduplication averaged over ten random seeds. We consider minority classes by gender, skin tone, and age. The difference between means of SemDeDup and \fairdedup across trials is significant at ${>}99.9\%$ confidence ($n{=}10$) for all groups according to a paired t-test. In all cases, we observe that \fairdedup helps recover mass reallocated to the majority class by SemDeDup.} 
\label{tab:facet-distribution}
\end{table}

\xhdr{Variants and Applications of the FairDedup Algorithm.} In our experiments, we use text-based prototypes to guide \fairdedup towards balancing representation of sensitive concepts. However, the exact specification of these prototypes is flexible to other subjects (\eg, non-person related) and modalities (\eg, image-based concepts). \fairdedup can be trivially modified to consider any prototype for which the embedding model can output similarity to individual images, such as sets of semantically aligned images (e.g., based on image type, photographs, illustrations, infographics, etc), or a combination of image and text prototypes. Similarly, FairDedup can be used to boost underrepresented samples from arbitrary sets such as object entities~\cite{xu2023demystifying},
or other forms of semantic organization.

\subsection{Limitations}
\label{sec:limitations}

\xhdrflat{Clustering Restrictions on Selection.}
While clustering allows deduplication algorithms to scale to hundreds of millions of samples, it also limits the availability of lower-represented samples for balancing sensitive concept representation. Take, for example, a data subset capturing photos of dancers. If the clustering algorithm creates two ``dancer'' clusters, bifurcating across binary gender presentation, then \fairdedup will be unable to perform significant gender balancing due to the independent processing of each cluster. We note that the resulting balance will be based on a combination of the underlying number of ``dancer'' photos in the dataset and the rate of duplication within both groups. If the two clusters are approximately equal sized with equal frequency of semantic duplicates, the independent deduplication of both clusters is equivalent in representation to a joint deduplication with respect to the bifurcated attribute. We display demonstrative clusters in the appendix (\cref{sec:sup-clustering}).

\xhdr{Bias Transfer From the Embedding Model.}
By deduplicating based on model embeddings, we subject the selection of samples to the biases of the embedding model. The majority of sensitive concepts we select are social constructs based in gender and race, and are not identifiable by anyone other than the photographed individual. We therefore expect sensitive concept representation to be based upon the predominate social norms they capture, rather than necessarily true identities of individuals. Nonetheless, we assert that a deduplication method which maintains the bias of the full-data setting is a favorable start to one that magnifies it.

\xhdr{Demographic Representation in Fairness Datasets.}
Most contemporary fairness datasets lack annotations from the individuals they represent. Consequently, for nonstationary socially constructed attributes such as gender, race, and perceived \emph{young/oldness}, the captured data relies solely on annotators' subjective understanding. Additionally, these datasets often limit gender representation to a binary perspective (occasionally including a small ``other'' category) \cite{devinney2022theories}, a necessary operationalization for scale that is not inclusive of bias characterization for diverse gender identities. We also note that fairness datasets cover a limited number of directions under which a model may express bias, excluding disability, national origin, and other sensitive attributes. Our analysis, therefore, only examines fairness outcomes with respect to contemporary and subjective evaluation of these limited available demographic attributes.

\section{Conclusion}
\label{sec:conclusion}
In this paper, we study the fairness outcomes resulting from training large-scale vision-language models on semantically deduplicated web-scale data, using LAION-400M and SemDeDup as a representative dataset and deduplication algorithm pairing. We find that deduplication has consistently harmful effects on gender-based bias and mixed effects on skin tone/race- and age-based biases across zero-shot classification and text-image retrieval tasks. To improve fairness outcomes, we propose \fairdedup, a simple and efficient fairness-aware modification of the sample selection heuristic in SemDeDup 
which boosts the representation of user-defined sensitive concepts in the post-deduplication data distribution. Our experiments show that \fairdedup preserves the performance of the full-data setting on standard metrics for common image-text datasets, has more favorable fairness outcomes than SemDeDup across all cases for gender- and skin tone/race-based biases, and outperforms the baseline full-data setting in several instances. We hope for \fairdedup to provide a simple and tractable baseline for future work in fairness-aware deduplication.

{
    \small
    \bibliographystyle{style/ieeenat_fullname}
    \bibliography{main}
}

\clearpage
\setcounter{page}{1}
\maketitlesupplementary

\section{Bias Constrained Clusters}
\label{sec:sup-clustering}

\begin{figure}[h]
\centering
\begin{subfigure}[b]{\linewidth}
\centering
\includegraphics[width=0.49\linewidth]{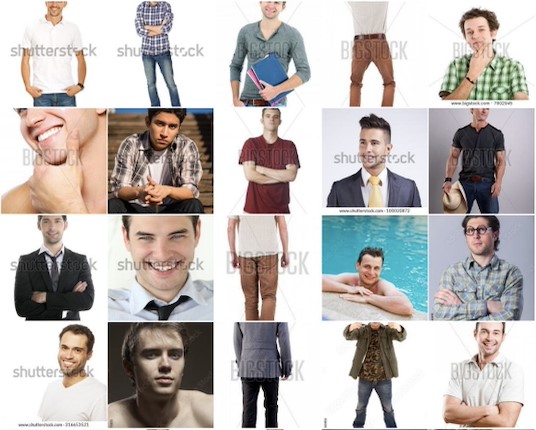}
\includegraphics[width=0.49\linewidth]{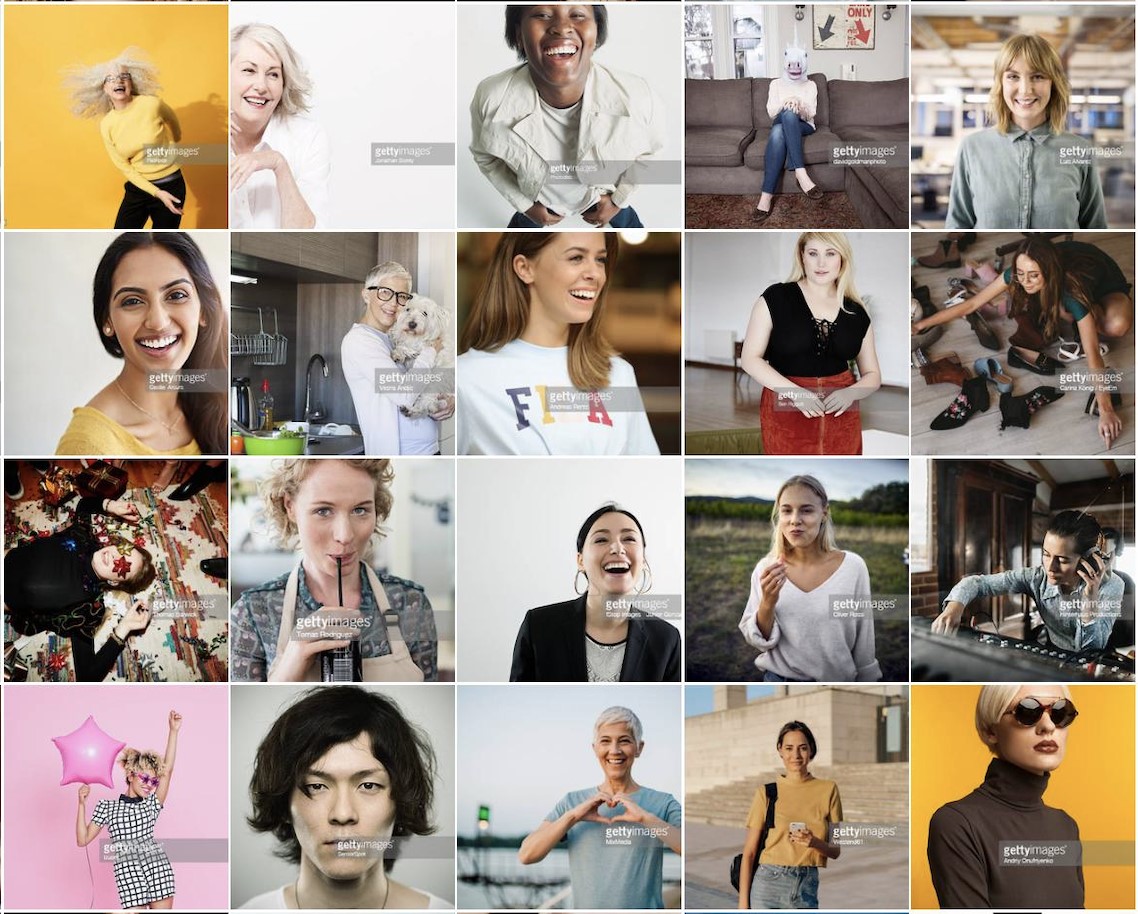}
\includegraphics[width=0.49\linewidth]{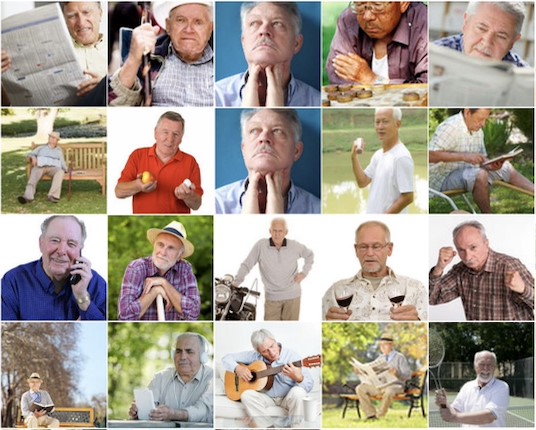}
\includegraphics[width=0.49\linewidth]{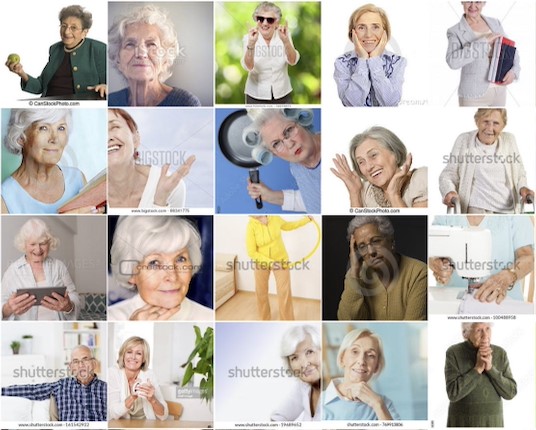}
\caption{Stock close-ups bifurcated across gender presentation and age.}
\end{subfigure}
\caption{Sample Clusters where selecting for certain underrepresented concepts may be difficult due to them being split into an entirely different cluster.}
\label{fig:sup-cluster}
\end{figure}

As described in \cref{sec:discussion}, clustering may limit the availability of lower represented samples for biasing sensitive concept representation. We show several demonstrative clusters in \cref{fig:sup-cluster} alongside descriptions of possible limitations.

\newpage

\section{Hyperparameters}
\label{sec:sup-hyperparameters}
\begin{table}[h]
\centering
\begin{tabular}{ll}
\toprule
Parameter                     & Value                            \\ \midrule
Model                         & CLIP~\cite{radford2021clip}      \\
Image Encoder                 & Vision Transformer Base/16 \cite{dosovitskiy2021vit} \\
Text Encoder                  & Text Transformer \cite{vaswani2017attention} \\
Epochs                        & 16                               \\
Batch Size                    & 33,820                           \\
Learning Rate                 & $5{\times}10^{-4}$               \\
LR Warmup                     & Linear / 2,000 Batches           \\
LR Schedule                   & Cosine Annealing                 \\
Optimizer                     & AdamW~\cite{loshchilov2017adamw} \\ 
\hspace{2em} $Decay$          & 0.2 \\
\hspace{2em} $\beta_1$        & 0.90 \\
\hspace{2em} $\beta_2$        & 0.98 \\
\hspace{2em} $\epsilon$       & $1{\times}10^{-6}$ \\
Precision                     & AMP BFloat16       \\
\bottomrule
\end{tabular}
\caption{Hyperparameters used to train all models.}
\label{tab:hyperparams-sup}
\end{table}

\section{Choosing Sensitive Concepts}
\label{sec:sup-concepts}
We derive sensitive concepts based on commonly protected groups in law. For example, Title VII of the US Civil Rights Act prohibits employment discrimination based on groups like \textit{race} and \textit{religion}. We take the intersection of commonly protected groups and those which are annotated in VL fairness datasets (\eg, only \textit{race} from above) to inform our concept design. In this way, we are able to consider concepts relevant to real-world practice while retaining our ability to evaluate the effectiveness of our mitigation strategy in a lab setting. To represent intersectional identities, we specifically choose sensitive concepts which simultaneously capture many of these protected groups.

\newpage
\onecolumn
\section{Deduplicated Cluster Examples}
\label{sec:sup-selection}
\begin{figure}[hbt!]
    \centering
    \begin{subfigure}[b]{\linewidth}
        \centering
        \includegraphics[width=0.36\linewidth]{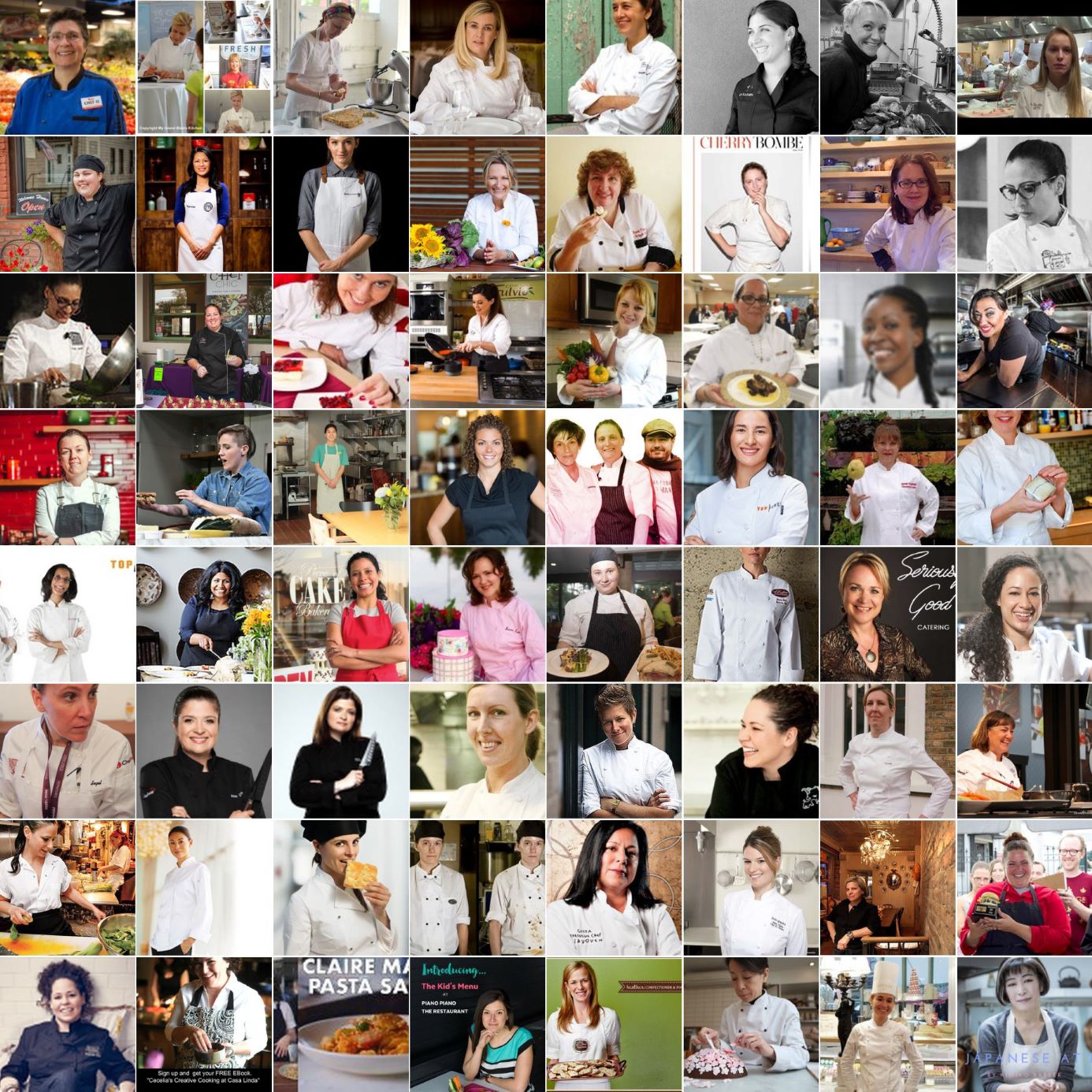}
        \includegraphics[width=0.36\linewidth]{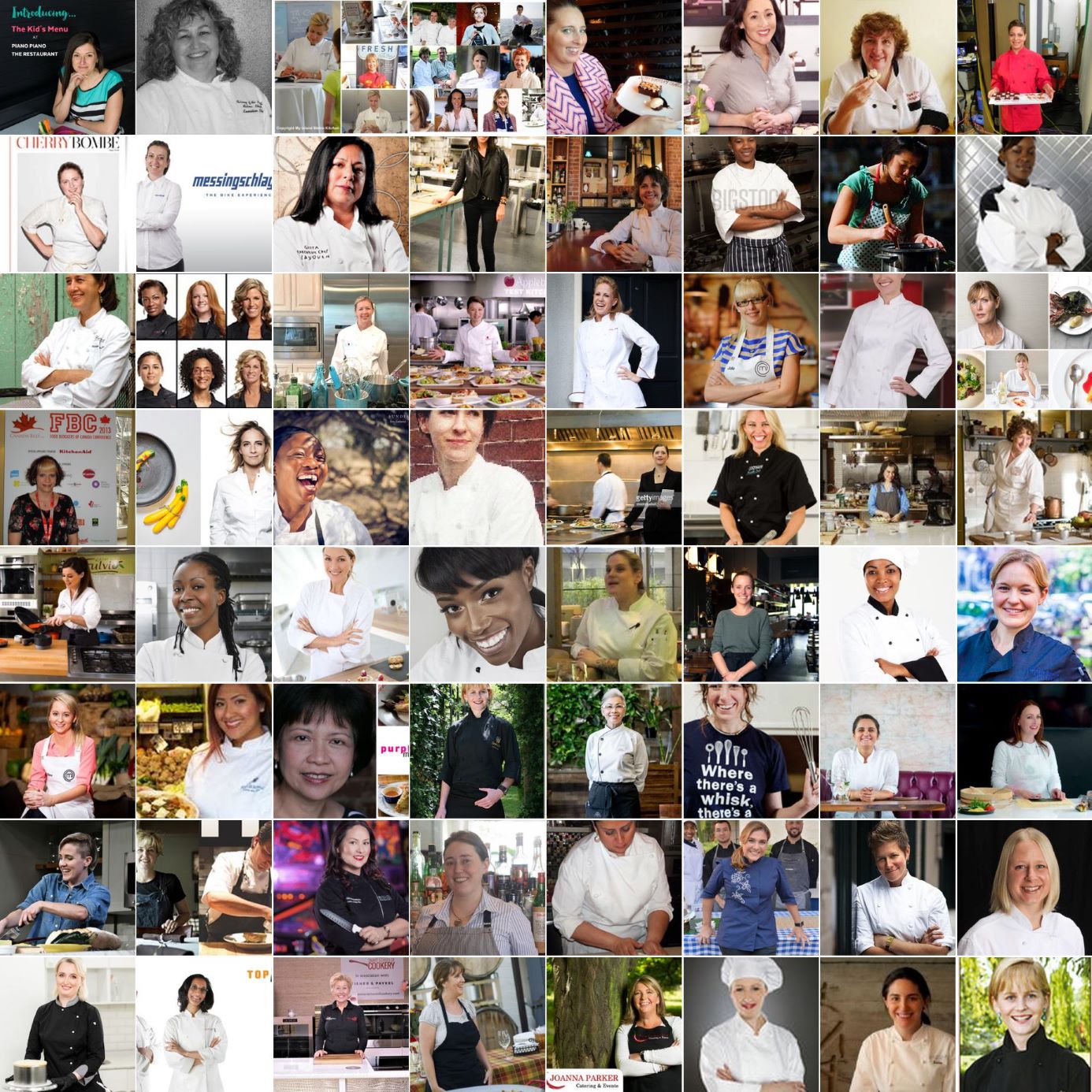}
        \caption{Chef}
    \end{subfigure}
    \begin{subfigure}[b]{\linewidth}
        \centering
        \includegraphics[width=0.36\linewidth]{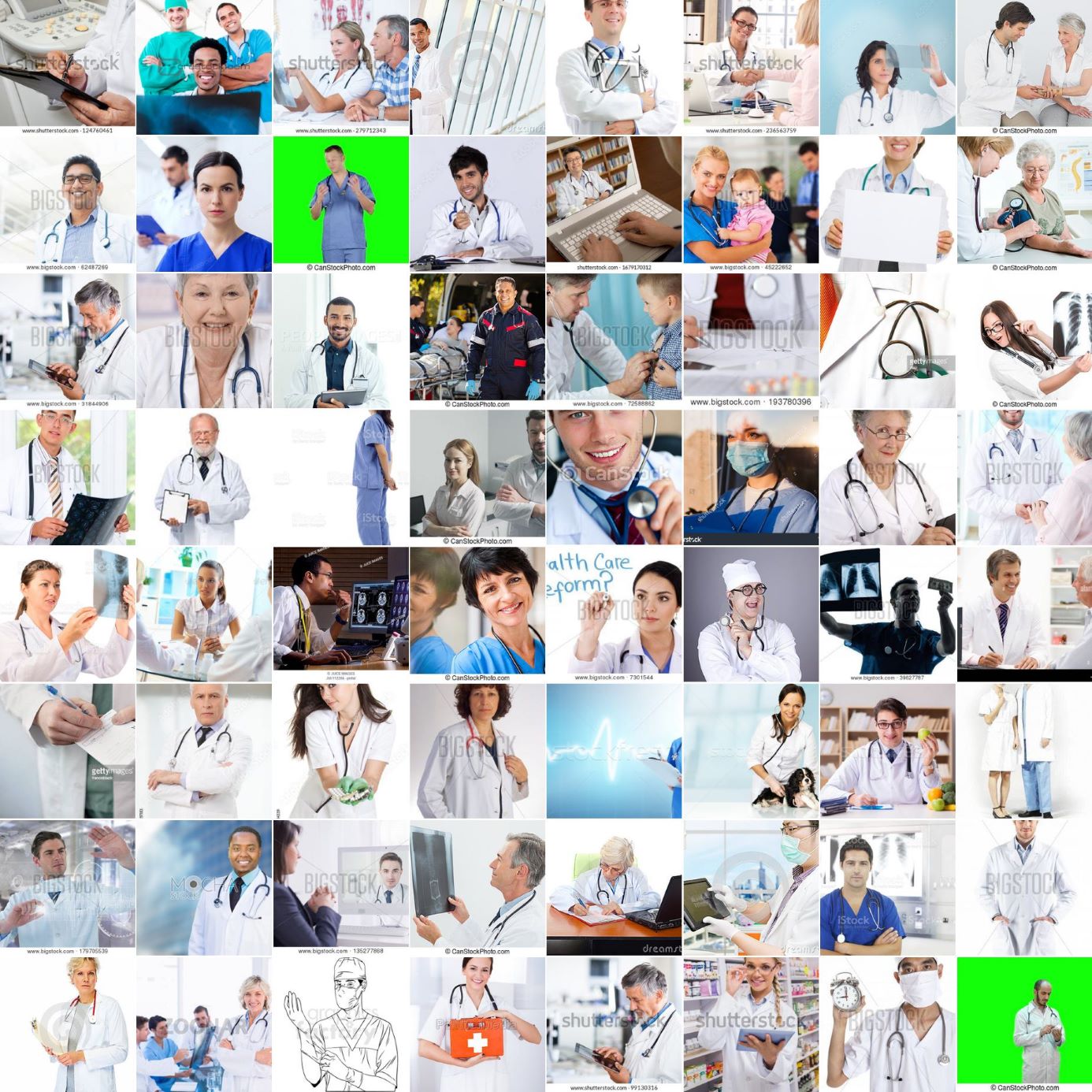}
        \includegraphics[width=0.36\linewidth]{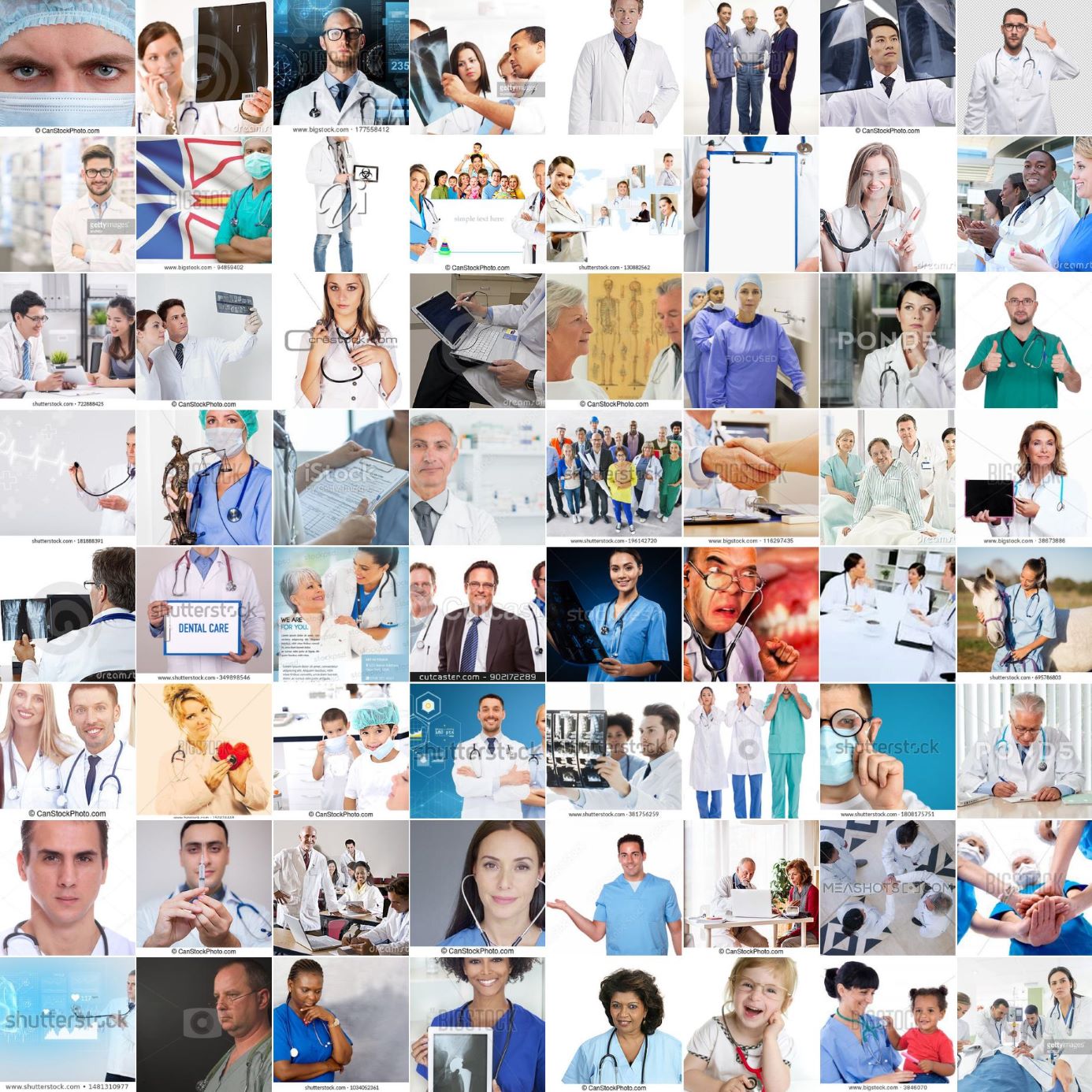}
        \caption{Doctor}
    \end{subfigure}
    \begin{subfigure}[b]{\linewidth}
        \centering
        \includegraphics[width=0.36\linewidth]{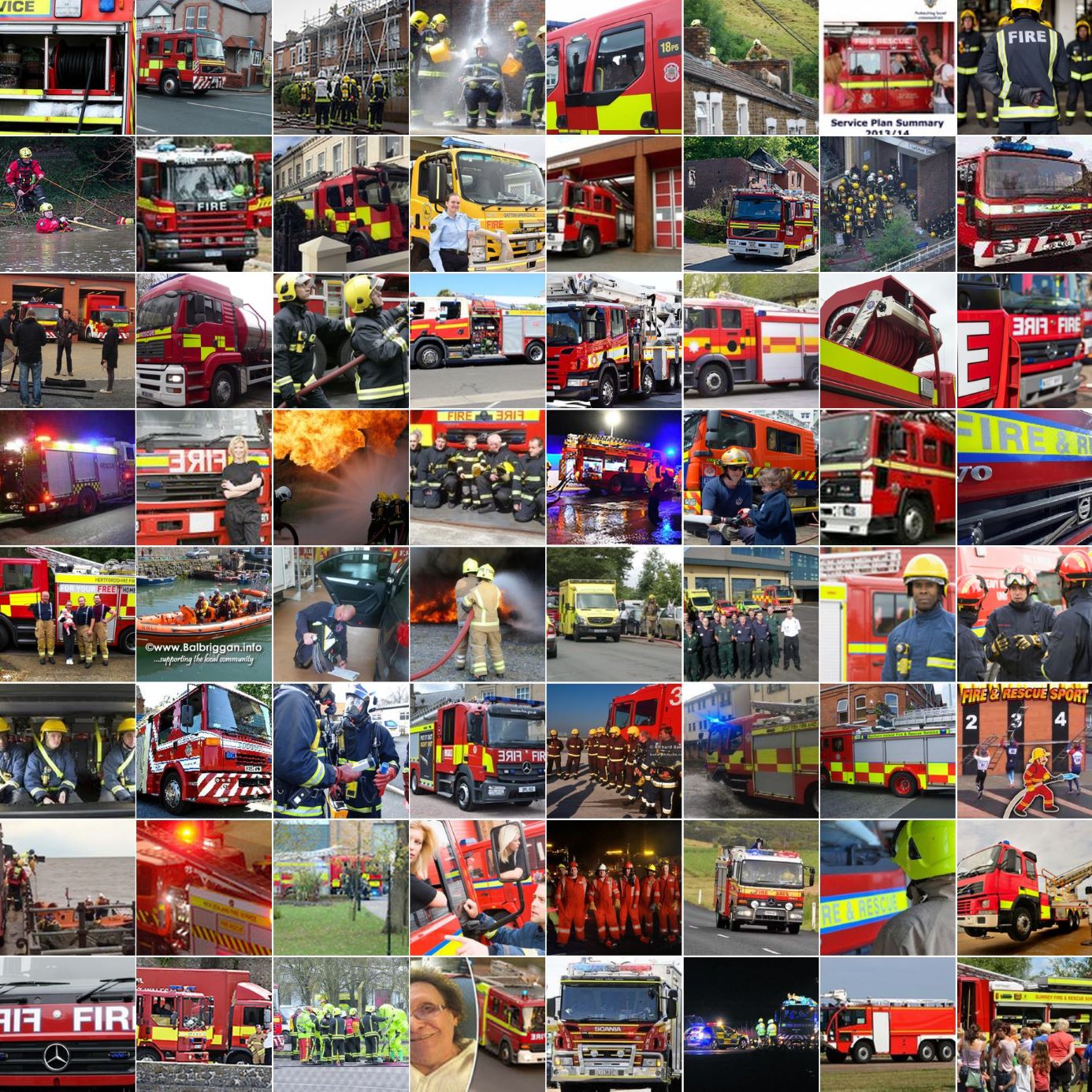}
        \includegraphics[width=0.36\linewidth]{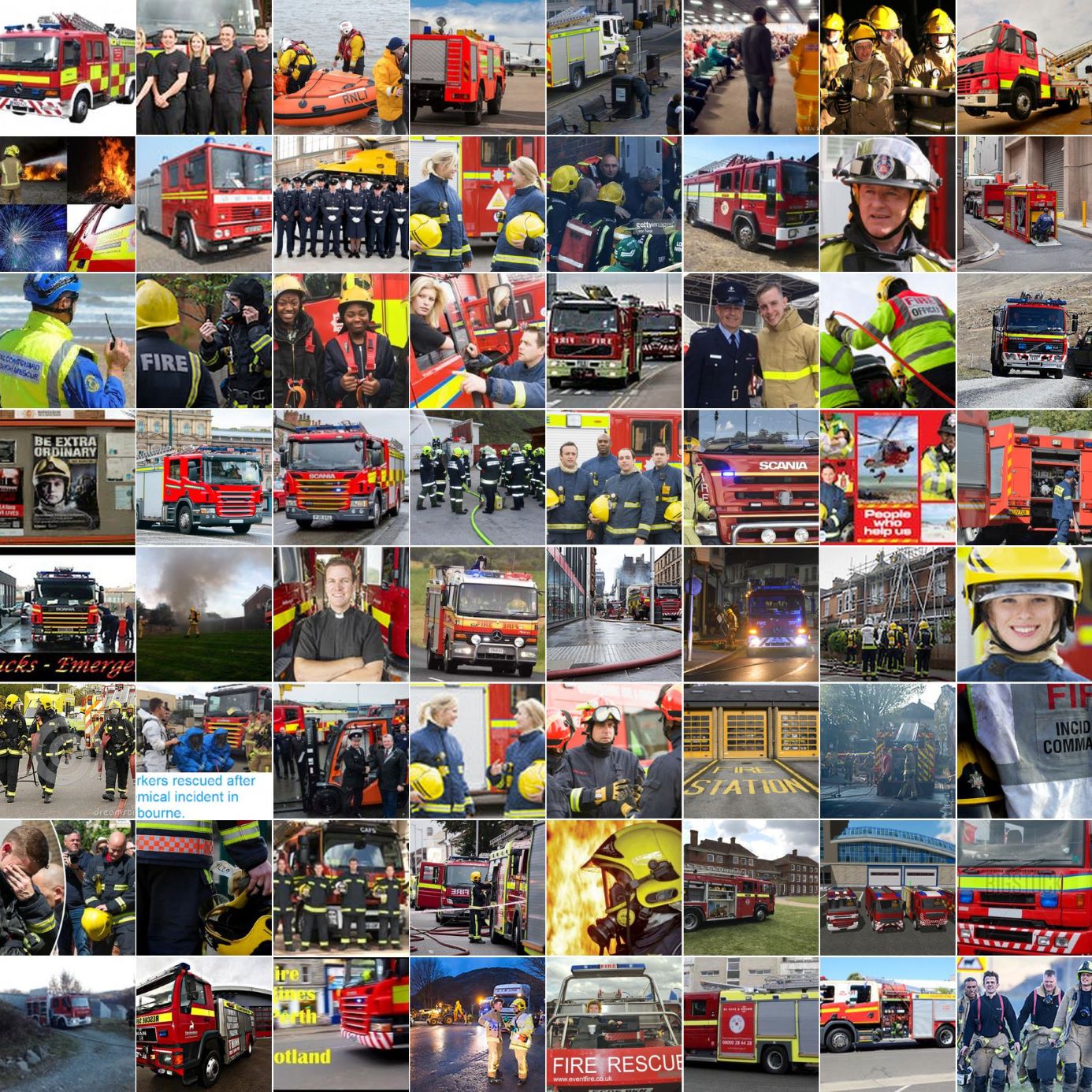}
        \caption{Firefighter}
    \end{subfigure}
    \caption{Additional samples from person-related clusters after applying \textbf{SemDeDup (left)} and \textbf{FairDeDup (right)}. Samples are randomly selected from clusters manually identified to include people. We annotate each cluster with our identification of its semantic contents.}
    \label{fig:sup-deduped}
\end{figure}

\newpage
\section{Sensitive Concept Prototypes}
\label{sec:sup-prototype}
\begin{figure}[hbt!]
\centering
\begin{subfigure}[b]{\linewidth}
\begin{subfigure}[b]{0.48\linewidth}
\centering
\begin{lstlisting}
person 
woman 
man 
black person 
black woman 
black man 
white person 
white woman 
white man 
indian person 
indian woman 
indian man 
latino person 
latino woman 
latino man 
east asian person 
east asian woman 
east asian man 
middle eastern person 
middle eastern woman 
middle eastern man 
southeast asian person 
southeast asian woman 
southeast asian man 
old person
old woman
old man
old black person
old black woman
old black man
old white person
old white woman
old white man
old indian person
old indian woman
old indian man
old latino person
old latino woman
old latino man
old east asian person
old east asian woman
old east asian man
old middle eastern person
old middle eastern woman
old middle eastern man
old southeast asian person
old southeast asian woman
old southeast asian man
young person
young woman
young man
young black person
young black woman
young black man
young white person
\end{lstlisting}
\end{subfigure}
\hfill
\begin{subfigure}[b]{0.48\linewidth}
\centering
\begin{lstlisting}[firstnumber=56]
young white woman
young white man
young indian person
young indian woman
young indian man
young latino person
young latino woman
young latino man
young east asian person
young east asian woman
young east asian man
young middle eastern person
young middle eastern woman
young middle eastern man
young southeast asian person
young southeast asian woman
young southeast asian man
child
black child
white child
indian child
latino child
east asian child
middle eastern child
southeast asian child
baby
black baby
white baby
indian baby
latino baby
east asian baby
middle eastern baby
southeast asian baby
boy
girl
black boy
black girl
white boy
white girl
indian boy
indian girl
latino boy
latino girl
east asian boy
east asian girl
middle eastern boy
middle eastern girl
southeast asian boy
southeast asian girl
person with dark skin
person with light skin
old person with dark skin
old person with light skin
young person with dark skin
young person with light skin
\end{lstlisting}
\end{subfigure}
\vspace{-5pt}
\caption{Sensitive Concepts}
\end{subfigure}
\begin{subfigure}[b]{\linewidth}
\begin{lstlisting}
A photo of a {concept}
This is a photo of a {concept}
A {concept}
\end{lstlisting}
\vspace{-5pt}
\caption{Text Templates}
\end{subfigure}
\vspace{-15pt}
\caption{Sensitive concepts and templates used to generate text concept prototypes for \fairdedup in \cref{sec:methodology}.}
\label{fig:prototype-sup}
\end{figure}

\twocolumn
\newpage
\section{Additional Datasets and Metrics}
\label{sec:sup-benchmarks}
\begin{table}[ht!]
\centering
\begin{tabular}{l@{\hspace{.25em}}l@{\hspace{.50em}}rccc}
\toprule
& & &  \makecell{\textbf{Full Data} \\[-5pt] {\tiny ($100\%$)}}
& \makecell{\textbf{SemDeDup}  \\[-5pt] {\tiny ($50\%$)}}
& \makecell{\textbf{\fairdedup} \\[-5pt] {\tiny ($50\%$)}} \\
\midrule
\multirow{3}{*}{\rcnt{\bf -a}} &
\multirow{3}{*}{\rcnt{\scriptsize}} &
Acc@1 & {.2947} & {.3096} & {.2988} \\
& & Acc@5 & {.6247} & {.6359} & {.6271} \\
& & MPCR & {.3163} & {.3228} & {.3126} \\
\midrule
\multirow{3}{*}{\rcnt{\bf -o}} &
\multirow{3}{*}{\rcnt{\scriptsize}} &
Acc@1 & {.5060} & {.5365} & {.5360} \\
& & Acc@5 & {.8235} & {.8420} & {.8385} \\
& & MPCR & {.5213} & {.5501} & {.5443} \\
\midrule
\multirow{3}{*}{\rcnt{\bf -r}} &
\multirow{3}{*}{\rcnt{\scriptsize}} &
Acc@1 & {.7646} & {.7652} & {.7534} \\
& & Acc@5 & {.9249} & {.9245} & {.9223} \\
& & MPCR & {.7512} & {.7506} & {.7386} \\
\midrule
\multirow{3}{*}{\rcnt{\bf 1k}} &
\multirow{3}{*}{\rcnt{\scriptsize}} &
Acc@1 & {.6640} & {.6579} & {.6535} \\
& & Acc@5 & {.8993} & {.8974} & {.8970} \\
& & MPCR & {.6639} & {.6578} & {.6536} \\
\midrule
\multirow{3}{*}{\rcnt{\bf sketch}} &
\multirow{3}{*}{\rcnt{\scriptsize }} &
Acc@1 & {.5142} & {.5036} & {.4989} \\
& & Acc@5 & {.7803} & {.7774} & {.7724} \\
& & MPCR & {.5144} & {.5040} & {.4993} \\
\midrule
\multirow{3}{*}{\rcnt{\bf v2}} &
\multirow{3}{*}{\rcnt{\scriptsize }} &
Acc@1 & {.5821} & {.5776} & {.5747} \\
& & Acc@5 & {.8453} & {.8413} & {.8368} \\
& & MPCR & {.5822} & {.5774} & {.5747} \\
\bottomrule
\end{tabular}
\caption{Zero-shot classification performance on ImageNet~\cite{deng2009imagenet} variants from CLIP Benchmark~\cite{cherti2023clipbenchmark}. Mean Per Class Recall abbreviated as MPCR. Higher is better for all metrics.}
\label{tab:benchmark-sup-imagenet}
\end{table}
\begin{table}[ht!]
\centering
\begin{tabular}{l@{\hspace{.25em}}l@{\hspace{.50em}}rccc}
\toprule
& & &  \makecell{\textbf{Full Data} \\[-5pt] {\tiny ($100\%$)}}
& \makecell{\textbf{SemDeDup}  \\[-5pt] {\tiny ($50\%$)}}
& \makecell{\textbf{\fairdedup} \\[-5pt] {\tiny ($50\%$)}} \\
\multirow{2}{*}{\rcnt{\bf \scriptsize flickr30k}} &
\multirow{2}{*}{\rcnt{\scriptsize}} &
   Image R@5 & {.8728} & {.8736} & {.8714} \\
& & Text R@5 & {.9640} & {.9670} & {.9620} \\
\midrule
\multirow{2}{*}{\rcnt{\bf \scriptsize flickr8k}} &
\multirow{2}{*}{\rcnt{\scriptsize}} &
   Image R@5 & {.8570} & {.8592} & {.8574} \\
& & Text R@5 & {.9450} & {.9400} & {.9300} \\
\midrule
\multirow{2}{*}{\rcnt{\bf \scriptsize coco}} &
\multirow{2}{*}{\rcnt{\scriptsize}} &
   Image R@5 & {.6325} & {.6318} & {.6255} \\
& & Text R@5 & {.7852} & {.7882} & {.7810} \\
\bottomrule
\end{tabular}
\caption{Image-text (Image R@5) and text-image (Text R@5) retrieval Recall@5 on Flickr~\cite{young2014flickr} and COCO~\cite{lin2014coco} from CLIP Benchmark~\cite{cherti2023clipbenchmark}. Higher is better for all metrics.}
\label{tab:benchmark-sup-imagenet-ret}
\end{table}
\begin{table}[ht!]
\centering
\begin{tabular}{l@{\hspace{.25em}}l@{\hspace{.50em}}rccc}
\toprule
& & &  \makecell{\textbf{Full Data} \\[-5pt] {\tiny ($100\%$)}}
& \makecell{\textbf{SemDeDup}  \\[-5pt] {\tiny ($50\%$)}}
& \makecell{\textbf{\fairdedup} \\[-5pt] {\tiny ($50\%$)}} \\
\midrule
\multirow{3}{*}{\rcnt{\bf cars}} &
\multirow{3}{*}{\rcnt{\scriptsize }} &
    Acc@1 & {.8526} & {.8346} & {.8429} \\
& & Acc@5 & {.9923} & {.9909} & {.9922} \\
& & MPCR & {.8541} & {.8344} & {.8429} \\
\midrule
\multirow{3}{*}{\rcnt{\bf country}} &
\multirow{3}{*}{\rcnt{\scriptsize 211}} &
    Acc@1 & {.1791} & {.1742} & {.1685} \\
& & Acc@5 & {.3978} & {.3950} & {.3889} \\
& & MPCR & {.1789} & {.1740} & {.1683} \\
\midrule
\multirow{3}{*}{\rcnt{\bf fer}} &
\multirow{3}{*}{\rcnt{\scriptsize 2013}} &
    Acc@1 & {.3976} & {.3969} & {.4809} \\
& & Acc@5 & {.9358} & {.9115} & {.9413} \\
& & MPCR & {.4057} & {.3702} & {.4152} \\
\midrule
\multirow{3}{*}{\rcnt{\bf fgvc}} &
\multirow{3}{*}{\rcnt{\scriptsize aircraft}} &
Acc@1 & {.1563} & {.1497} & {.1665} \\
& & Acc@5 & {.4164} & {.4146} & {.4212} \\
& & MPCR & {.1545} & {.1495} & {.1667} \\
\midrule
\multirow{3}{*}{\rcnt{\bf gtsrb}} &
\multirow{3}{*}{\rcnt{\scriptsize }} &
Acc@1 & {.4159} & {.4101} & {.3383} \\
& & Acc@5 & {.7352} & {.6814} & {.7062} \\
& & MPCR & {.3884} & {.3806} & {.3571} \\
\midrule
\multirow{3}{*}{\rcnt{\bf mnist}} &
\multirow{3}{*}{\rcnt{\scriptsize mnist}} &
Acc@1 & {.3896} & {.5474} & {.5890} \\
& & Acc@5 & {.8064} & {.8022} & {.9058} \\
& & MPCR & {.3938} & {.5531} & {.5911} \\
\midrule
\multirow{3}{*}{\rcnt{\bf objectnet}} &
\multirow{3}{*}{\rcnt{\scriptsize }} &
Acc@1 & {.4722} & {.4754} & {.4781} \\
& & Acc@5 & {.7296} & {.7283} & {.7311} \\
& & MPCR & {.4614} & {.4681} & {.4665} \\
\midrule
\multirow{2}{*}{\rcnt{\bf render}} &
\multirow{2}{*}{\rcnt{\scriptsize sst2}} &
Acc@1 & {.5371} & {.5157} & {.5041} \\
& & MPCR & {.5368} & {.5155} & {.5036} \\
\midrule
\multirow{3}{*}{\rcnt{\bf stl10}} &
\multirow{3}{*}{\rcnt{\scriptsize }} &
Acc@1 & {.9659} & {.9705} & {.9686} \\
& & Acc@5 & {.9996} & {.9998} & {.9995} \\
& & MPCR & {.9659} & {.9708} & {.9684} \\
\midrule
\multirow{3}{*}{\rcnt{\bf sun397}} &
\multirow{3}{*}{\rcnt{\scriptsize }} &
Acc@1 & {.6856} & {.6786} & {.6838} \\
& & Acc@5 & {.9344} & {.9351} & {.9363} \\
& & MPCR & {.6730} & {.6613} & {.6662} \\
\midrule
\multirow{3}{*}{\rcnt{\bf voc2007}} &
\multirow{3}{*}{\rcnt{\scriptsize }} &
Acc@1 & {.7432} & {.7617} & {.7584} \\
& & Acc@5 & {.9498} & {.9518} & {.9515} \\
& & MPCR & {.8197} & {.8297} & {.8286} \\
\bottomrule
\end{tabular}
\caption{Additional zero-shot classification results from CLIP Benchmark~\cite{cherti2023clipbenchmark}. Mean Per Class Recall abbreivated as MPCR. Higher is better for all metrics.}
\label{tab:benchmark-sup-cla}
\end{table}

\makeatletter
\setlength{\@fptop}{0pt}
\makeatother
\begin{table}[ht!]
\centering
\begin{tabular}{l@{\hspace{.25em}}l@{\hspace{.50em}}rccc}
\toprule
& & &  \makecell{\textbf{Full Data} \\[-5pt] {\tiny ($100\%$)}}
& \makecell{\textbf{SemDeDup}  \\[-5pt] {\tiny ($50\%$)}}
& \makecell{\textbf{\fairdedup} \\[-5pt] {\tiny ($50\%$)}} \\
\midrule
\multirow{3}{*}{\rcnt{\bf caltech}} &
\multirow{3}{*}{\rcnt{\scriptsize 101}} &
Acc@1 & {.8304} & {.8230} & {.8309} \\
& & Acc@5 & {.9399} & {.9394} & {.9578} \\
& & MPCR & {.9193} & {.9035} & {.9061} \\
\midrule
\multirow{3}{*}{\rcnt{\bf cifar}} &
\multirow{3}{*}{\rcnt{\scriptsize 10}} &
Acc@1 & {.9198} & {.9255} & {.9203} \\
& & Acc@5 & {.9986} & {.9984} & {.9992} \\
& & MPCR & {.9198} & {.9254} & {.9204} \\
\midrule
\multirow{3}{*}{\rcnt{\bf cifar}} &
\multirow{3}{*}{\rcnt{\scriptsize 100}} &
Acc@1 & {.7234} & {.7291} & {.7299} \\
& & Acc@5 & {.9342} & {.9342} & {.9386} \\
& & MPCR & {.7231} & {.7289} & {.7303} \\
\midrule
\multirow{3}{*}{\rcnt{\bf clevr}} &
\multirow{3}{*}{\rcnt{\scriptsize obj dist}} &
Acc@1 & {.2033} & {.2101} & {.2232} \\
& & Acc@5 & {.9187} & {.9187} & {.9187} \\
& & MPCR & {.1686} & {.1633} & {.1673} \\
\midrule
\multirow{3}{*}{\rcnt{\bf clevr}} &
\multirow{3}{*}{\rcnt{\scriptsize count all}} &
Acc@1 & {.1421} & {.2317} & {.1916} \\
& & Acc@5 & {.6474} & {.8174} & {.6397} \\
& & MPCR & {.1397} & {.2264} & {.1873} \\
\midrule
\multirow{3}{*}{\rcnt{\bf diabetic}} &
\multirow{3}{*}{\rcnt{\scriptsize }} &
Acc@1 & {.0646} & {.3209} & {.0666} \\
& & Acc@5 & {1.0000} & {1.0000} & {1.0000} \\
& & MPCR & {.2178} & {.1998} & {.2029} \\
\midrule
\multirow{3}{*}{\rcnt{\bf dmlab}} &
\multirow{3}{*}{\rcnt{\scriptsize }} &
Acc@1 & {.1949} & {.1692} & {.1960} \\
& & Acc@5 & {.8430} & {.8236} & {.8270} \\
& & MPCR & {.1677} & {.1869} & {.1569} \\
\midrule
\multirow{3}{*}{\rcnt{\bf dsprites}} &
\multirow{3}{*}{\rcnt{\scriptsize label orient.}} &
Acc@1 & {.0226} & {.0272} & {.0244} \\
& & Acc@5 & {.1259} & {.1264} & {.1302} \\
& & MPCR & {.0231} & {.0276} & {.0249} \\
\midrule
\multirow{3}{*}{\rcnt{\bf dsprites}} &
\multirow{3}{*}{\rcnt{\scriptsize label xpos}} &
Acc@1 & {.0305} & {.0315} & {.0305} \\
& & Acc@5 & {.1600} & {.1568} & {.1625} \\
& & MPCR & {.0313} & {.0321} & {.0312} \\
\midrule
\multirow{3}{*}{\rcnt{\bf dsprites}} &
\multirow{3}{*}{\rcnt{\scriptsize label ypos}} &
Acc@1 & {.0315} & {.0317} & {.0317} \\
& & Acc@5 & {.1553} & {.1559} & {.1596} \\
& & MPCR & {.0311} & {.0312} & {.0312} \\
\bottomrule
\end{tabular}
\caption{Zero-shot classification performance on Visual Task Adaptation Benchmark (VTAB)~\cite{zhai2019vtab} datasets from CLIP Benchmark~\cite{cherti2023clipbenchmark}. Mean Per Class Recall abbreviated as MPCR. Higher is better for all metrics.}
\label{tab:benchmark-sup-vtab1}
\end{table}

\newpage

\begin{table}[ht!]
\centering
\begin{tabular}{l@{\hspace{.25em}}l@{\hspace{.50em}}rccc}
\toprule
& & &  \makecell{\textbf{Full Data} \\[-5pt] {\tiny ($100\%$)}}
& \makecell{\textbf{SemDeDup}  \\[-5pt] {\tiny ($50\%$)}}
& \makecell{\textbf{\fairdedup} \\[-5pt] {\tiny ($50\%$)}} \\
\midrule
\multirow{3}{*}{\rcnt{\bf dtd}} &
\multirow{3}{*}{\rcnt{\scriptsize }} &
Acc@1 & {.5330} & {.5021} & {.5074} \\
& & Acc@5 & {.8293} & {.7883} & {.8202} \\
& & MPCR & {.5330} & {.5011} & {.5080} \\
\midrule
\multirow{3}{*}{\rcnt{\bf eurosat}} &
\multirow{3}{*}{\rcnt{\scriptsize }} &
Acc@1 & {.4904} & {.5220} & {.5246} \\
& & Acc@5 & {.9335} & {.9289} & {.8987} \\
& & MPCR & {.5117} & {.5288} & {.5404} \\
\midrule
\multirow{3}{*}{\rcnt{\bf flowers}} &
\multirow{3}{*}{\rcnt{\scriptsize }} &
Acc@1 & {.6723} & {.6536} & {.6878} \\
& & Acc@5 & {.8533} & {.8442} & {.8374} \\
& & MPCR & {.6657} & {.6384} & {.6509} \\
\midrule
\multirow{2}{*}{\rcnt{\bf kitti}} &
\multirow{2}{*}{\rcnt{\scriptsize dist.}} &
Acc@1 & {.1589} & {.1505} & {.2363} \\
& & MPCR & {.2134} & {.1707} & {.2134} \\
\midrule
\multirow{2}{*}{\rcnt{\bf pcam}} &
\multirow{2}{*}{\rcnt{\scriptsize }} &
Acc@1 & {.5521} & {.4720} & {.5910} \\
& & MPCR & {.5521} & {.4719} & {.5909} \\
\midrule
\multirow{3}{*}{\rcnt{\bf pets}} &
\multirow{3}{*}{\rcnt{\scriptsize }} &
Acc@1 & {.8929} & {.8577} & {.8812} \\
& & Acc@5 & {.9951} & {.9940} & {.9937} \\
& & MPCR & {.8917} & {.8570} & {.8802} \\
\midrule
\multirow{3}{*}{\rcnt{\bf resisc}} &
\multirow{3}{*}{\rcnt{\scriptsize 45}} &
Acc@1 & {.5694} & {.5892} & {.5763} \\
& & Acc@5 & {.8889} & {.8938} & {.8914} \\
& & MPCR & {.5751} & {.5955} & {.5853} \\
\midrule
\multirow{3}{*}{\rcnt{\bf smallnorb}} &
\multirow{3}{*}{\rcnt{\scriptsize label azimuth}} &
Acc@1 & {.0539} & {.0502} & {.0550} \\
& & Acc@5 & {.2796} & {.2781} & {.2738} \\
& & MPCR & {.0547} & {.0509} & {.0560} \\
\midrule
\multirow{3}{*}{\rcnt{\bf smallnorb}} &
\multirow{3}{*}{\rcnt{\scriptsize label elevation}} &
Acc@1 & {.0927} & {.1125} & {.1114} \\
& & Acc@5 & {.5319} & {.5770} & {.5544} \\
& & MPCR & {.0922} & {.1116} & {.1106} \\
\midrule
\multirow{3}{*}{\rcnt{\bf svhn}} &
\multirow{3}{*}{\rcnt{\scriptsize }} &
Acc@1 & {.3973} & {.3657} & {.4008} \\
& & Acc@5 & {.7782} & {.7694} & {.8107} \\
& & MPCR & {.3727} & {.3670} & {.3238} \\
\bottomrule
\end{tabular}
\caption{Additional results for VTAB~\cite{zhai2019vtab} extending \cref{tab:benchmark-sup-vtab1}.}
\label{tab:benchmark-sup-vtab2}
\end{table}

\cleardoublepage
\section{Extended Pseudo-Code}
\label{sec:sup-code}
\begin{figure}[h]
\centering
\begin{lstlisting}[language=PyTorch,escapechar=\%]
def semdedup(embs, eps):
    # Sort by distance to the cluster centroid.
    sort_by_dist_to_centroid(embs, desc=True)

    # Compute the pairwise cosine similarity
    pair_sims = embs @ embs.T
    triu_sims = torch.triu(pair_sims, diagonal=1)
    M = torch.max(triu_sim_matrix, dim=0)[0]

    # Keep if the max similarity <= threshold
    points = M <= 1 - epsilon
    log_and_keep(points)

def fairdedup(embs, prototypes, eps):
    # Get similarity with concept prototypes
    proto = embs @ prototypes.T
    
    balance = AverageMeter(prototype.shape[0])
    tovisit = torch.ones(embs.shape[0])
    while tovist.any():
        # Find an unvisited neighborhood
        node = torch.where(tovisit)[0][0]
        sims = embs[node] @ embs.T
        neighbors = torch.where(sims > 1 - eps)
        neighbors = neighbors[0]
    
        # Maximize least represented concept
        c = balance.get_min_concept()
        point = proto[neigbors][:, c].argmax()
        balance.update(point)
    
        log_and_keep(point)
        tovisit[neighbors] = 0

# Input: embedding_model, dataset, eps, prototypes
# Embed and cluster the dataset
embeddings = embedding_model(dataset)
per_cluster_embeddings = kmeans(embeddings)
for cluster in per_cluster_embeddings:
    # Choose selection method
    semdedup(cluster)
    fairdedup(cluster)
\end{lstlisting}
\caption{Extended PyTorch-style pseudo-code for SemDeDup and \fairdedup selection given concept \texttt{prototypes}, an \texttt{embedding\_model}, target \texttt{dataset} to deduplicate, and an \texttt{eps} similarity threshold for determining duplicates. SemDeDup pseudo-code modified from \cite{abbas2023semdedup}.}
\label{fig:pseudocode-sup}
\end{figure}

\end{document}